\documentclass[10pt,journal,compsoc]{IEEEtran}

\usepackage{color}
\usepackage{amsmath}
\usepackage{diagbox}
\usepackage{graphics}
\usepackage{epsfig}
\usepackage{threeparttable}
\usepackage{xspace}
\usepackage{siunitx}
\usepackage{graphicx}
\usepackage{amssymb}
\usepackage{threeparttable}
\usepackage{color}
\usepackage[normalem]{ulem}
\usepackage{tabularray} % 使用新的、更强大的表格宏包
\usepackage{arydshln} 
\usepackage{multirow}
\usepackage{float}
\usepackage{amsfonts}
\usepackage{bm}
\usepackage{bbold}
\usepackage{array}
\usepackage[table]{xcolor}
\usepackage{colortbl}
\usepackage{diagbox}
\usepackage{rotating}
\usepackage{booktabs}
\usepackage{overpic}
\usepackage{enumitem}
\usepackage{colortbl}
\usepackage{algorithm}
\usepackage{algorithmic}
\usepackage{cite}
\usepackage{dblfloatfix}
\usepackage[export]{adjustbox}
\usepackage{pifont}
\usepackage[pagebackref=false,breaklinks=true,colorlinks,bookmarks=false]{hyperref}
\usepackage{cleveref}
\usepackage{tikz}
\usepackage[edges]{forest}
\usepackage{caption} 
\usepackage{utfsym}
\definecolor{cvprblue}{rgb}{0.21,0.49,0.74}
\definecolor{nvidiagreen}{RGB}{119,185,0}
\usepackage{xcolor}  % 用于颜色
\usepackage{pifont}  % 用于 ding 符号
\usepackage{ragged2e}
\newcommand*{\belowrulesepcolor}[1]{% 
  \noalign{% 
    \kern-\belowrulesep 
    \begingroup 
      \color{#1}% 
      \hrule height\belowrulesep 
    \endgroup 
    \vspace{-0.03mm}
  }%
} 
\newcommand*{\aboverulesepcolor}[1]{% 
  \noalign{% 
  \vspace{-0.03mm}
    \begingroup 
      \color{#1}% 
      \hrule height\aboverulesep 
    \endgroup 
    \kern-\aboverulesep 
  }%
}

\usepackage{algorithm}
\usepackage{listings}
\usepackage{etoolbox}
\usepackage{hyperref}
\hypersetup{
    colorlinks=true,   % 将链接从边框改为彩色文本
    urlcolor=blue      % 设置URL链接的颜色为蓝色
}
\makeatletter
\AfterEndEnvironment{algorithm}{\let\@algcomment\relax}
\AtEndEnvironment{algorithm}{\kern2pt\hrule\relax\vskip3pt\@algcomment}
\let\@algcomment\relax
\newcommand\algcomment[1]{\def\@algcomment{\footnotesize#1}}
\renewcommand\fs@ruled{\def\@fs@cfont{\bfseries}\let\@fs@capt\floatc@ruled
  \def\@fs@pre{\hrule height.8pt depth0pt \kern2pt}%
  \def\@fs@post{}%
  \def\@fs@mid{\kern2pt\hrule\kern2pt}%
  \let\@fs@iftopcapt\iftrue}
\makeatother

\usepackage{tabulary}
\newcolumntype{I}{!{\vrule width 1pt}}
\newcolumntype{x}[1]{>{\centering\arraybackslash}p{#1pt}}
\newcolumntype{y}[1]{>{\raggedright\arraybackslash}p{#1pt}}
\newcolumntype{z}[1]{>{\raggedleft\arraybackslash}p{#1pt}}

\definecolor{mygray}{gray}{.85}
\definecolor{mygray1}{gray}{.7}
\definecolor{mygray2}{gray}{.93}
\definecolor{codegreen}{RGB}{79,126,127}
\definecolor{codedefine}{RGB}{153,54,159}
\definecolor{codefunc}{RGB}{73,122,234}
\definecolor{codecall}{RGB}{73,122,234}
\definecolor{codepro}{RGB}{212,96,80}
\definecolor{codedim}{RGB}{89,152,195}
\definecolor{3dgc1}{RGB}{177, 83, 74}
\definecolor{3dgc2}{RGB}{93, 107, 72}
\definecolor{hidden-draw}{RGB}{202,203,207}
\definecolor{hidden-2d}{RGB}{243,250,244}
\definecolor{hidden-video}{RGB}{250,248,243}
\definecolor{hidden-3d}{RGB}{236,249,253}
\definecolor{hidden-4d}{RGB}{244,243,250}

\makeatletter
\newcommand{\thickhline}{%
    \noalign {\ifnum 0=`}\fi \hrule height 1pt
    \futurelet \reserved@a \@xhline
}
\makeatother

\makeatletter
\DeclareRobustCommand\onedot{\futurelet\@let@token\@onedot}
\def\@onedot{\ifx\@let@token.\else.\null\fi\xspace}

\def\eg{\emph{e.g}\onedot} 
\def\ie{\emph{i.e}\onedot}

\def\etal{\emph{et al}\onedot}

\makeatother

\newcommand{\reffig}[1]{Fig.~\ref{#1}}
\newcommand{\reftab}[1]{Table~\ref{#1}}
\newcommand{\refsection}[1]{Section~\ref{#1}}

\begin{document}
\title{How Far are VLMs from Visual Spatial Intelligence? A Benchmark-Driven Perspective}
% How Far are VLMs from True Visual Spatial Intelligence? A Benchmark-Driven Perspective
% Advances in Visual Spatial Reasoning: A Survey
% \author{Songsong Yu$^{1,2}$$^\ast$, Yuxin Chen$^2$$^\ast$, Hao Ju$^3$$^\ast$, Lianjie Jia$^4$$^\ast$, Fuxi Zhang$^4$, 
% Shaofei Huang$^3$, Yuhan Wu$^4$, 
% \\Rundi Cui$^4$, Binghao Ran$^4$, Zaibin Zhang$^4$, Zhipeng Zhang$^1$, Yifan Wang$^4$, Lin Song$^2$, Zhedong Zheng$^3$,
% \\Lijun Wang$^4$, Yanwei Li$^5$$^\dag$, Ying Shan$^2$, Huchuan Lu$^4$,~\IEEEmembership{Fellow,~IEEE
% \\{\small $^1$SJTU, $^2$ARC Lab, Tencent PCG, $^3$UM, $^4$DLUT, $^5$CUHK}
% }
\author{Songsong Yu$^\ast$, Yuxin Chen$^\ast$, Hao Ju$^\ast$, Lianjie Jia$^\ast$, Fuxi Zhang, 
Shaofei Huang, Yuhan Wu, 
\\Rundi Cui, Binghao Ran, Zaibin Zhang, Zhipeng Zhang, Yifan Wang, Lin Song, Zhedong Zheng,
\\Lijun Wang, Yanwei Li$^\dag$, Ying Shan, Huchuan Lu,~\IEEEmembership{Fellow,~IEEE
% \\{\small $^1$SJTU, $^2$ARC Lab, Tencent PCG, $^3$UM, $^4$DLUT, $^5$CUHK}
}

\IEEEcompsocitemizethanks{
\IEEEcompsocthanksitem{$\ast$ Equal contribution, $^\dag$ Corresponding author: Yanwei Li. Work is done during the internship at Tencent ARCLab.}
\IEEEcompsocthanksitem {Songsong Yu is with Shanghai Jiao Tong University and an intern at ARC Lab, Tencent PCG.}
\IEEEcompsocthanksitem {Yuxin Chen, Lin Song, and Ying Shan are with ARC Lab, Tencent PCG.}
\IEEEcompsocthanksitem {Hao Ju, Zhedong Zheng, and Shaofei Huang are with the University of Macau, Macau SAR, China.}
\IEEEcompsocthanksitem {Lianjie Jia, Fuxi Zhang, Yuhan Wu, Rundi Cui, Binghao Ran, Zaibin Zhang, Yifan Wang, Lijun Wang, and Huchuan Lu are with Dalian University of Technology, China.}
\IEEEcompsocthanksitem {Zhipeng Zhang is with Shanghai Jiao Tong University, China.}
\IEEEcompsocthanksitem {Yanwei Li is with The Chinese University of Hong Kong, Hong Kong SAR, China. E-mail: liyanwei@link.cuhk.edu.hk}
}
}

% The paper headers
% \markboth{IEEE TRANSACTIONS ON PATTERN ANALYSIS AND MACHINE INTELLIGENCE}%
% {Shell \MakeLowercase{\textit{et al.}}: Bare Demo of IEEEtran.cls for Journals}

\newcommand{\myPara}[1]{\vspace{.05in}\noindent\textbf{#1}}
\IEEEtitleabstractindextext{
\justify
\begin{abstract}
Visual Spatial Reasoning (VSR) is a core human cognitive ability and a critical requirement for advancing embodied intelligence and autonomous systems. Despite recent progress in Vision-Language Models (VLMs), achieving human-level VSR remains highly challenging due to the complexity of representing and reasoning over three-dimensional space. In this paper, we present a systematic investigation of VSR in VLMs, encompassing a review of existing methodologies across input modalities, model architectures, training strategies, and reasoning mechanisms.
Furthermore, we categorize spatial intelligence into three levels of capability, \ie, basic perception, spatial understanding, spatial planning, and curate SIBench, a spatial intelligence benchmark encompassing nearly 20 open-source datasets across 23 task settings.
Experiments with state-of-the-art VLMs reveal a pronounced gap between perception and reasoning, as models show competence in basic perceptual tasks but consistently underperform in 
understanding and planning tasks, particularly in numerical estimation, multi-view reasoning, temporal dynamics, and spatial imagination. These findings underscore the substantial challenges that remain in achieving spatial intelligence, while providing both a systematic roadmap and a comprehensive benchmark to drive future research in the field.
The related resources of this study are accessible at 
\href{https://sibench.github.io/Awesome-Visual-Spatial-Reasoning/}{https://sibench.github.io/Awesome-Visual-Spatial-Reasoning/}.

% project page
\end{abstract}

}

\maketitle
%trends and
\IEEEdisplaynontitleabstractindextext
\IEEEpeerreviewmaketitle

\section{Introduction}
\begin{figure*}[!t]
\centering
\includegraphics[width=7.2in]{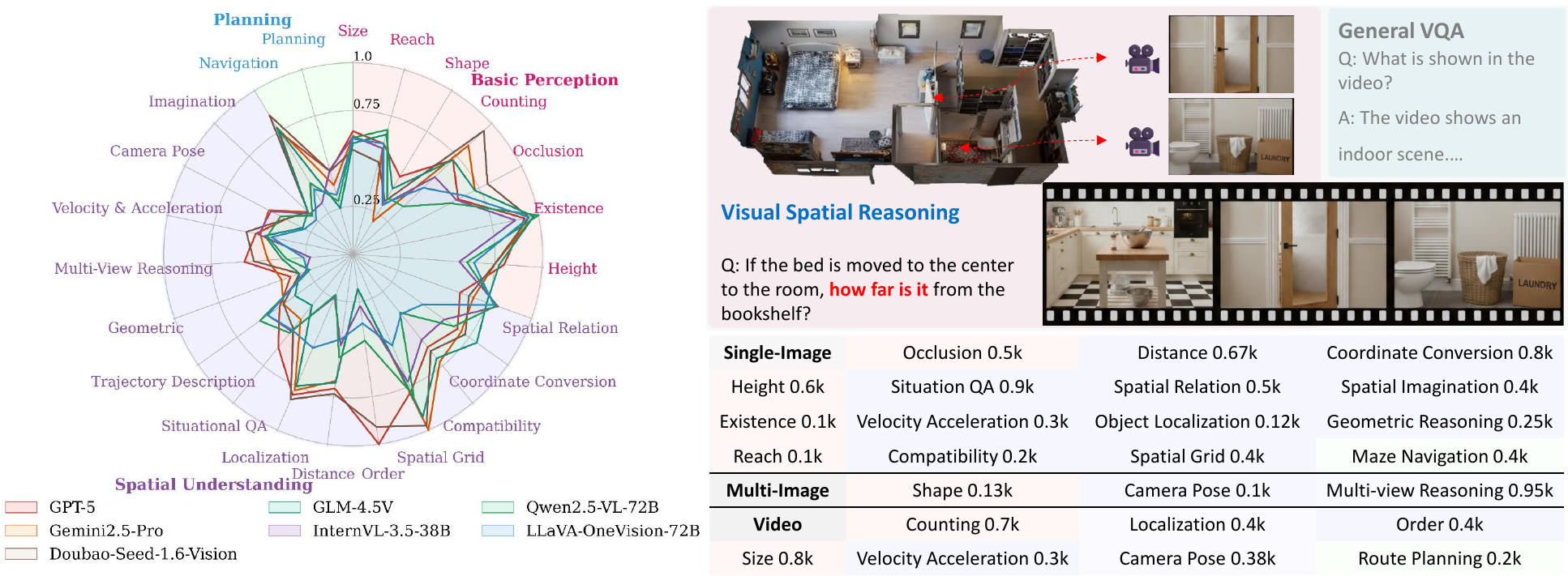}
\caption{
\textbf{Performance of SOTA Models on 23 Visual Spatial Reasoning Tasks} (left). The evaluation reveals that the models have significant room for improvement, especially in tasks requiring precise numerical estimation, perspective taking, temporal information processing, and, particularly, spatial imagination. See \reftab{tab:model_performance} and \reftab{tab:model_performance-mini} for detailed results.
\textbf{Comparison of Visual Spatial Reasoning and General VQA}~(Upper-right). While general VQA tasks primarily focus on extracting semantic information from images, VSR necessitates a deeper capacity to model and reason about spatial relationships.
\textbf{Data Formats and Task Settings for Visual Spatial Reasoning}~(Bottom-right). The evaluation includes 3 input formats and 23 task settings, covering three levels: Basic Perception, Spatial Understanding, and Planning.
}
% reach, order, realative-depth:SPAR-Bench, Compatibility spatial imagination Qwen-existence&occlusion 需要重新测
\label{fig:radar}
\end{figure*}

\IEEEPARstart{V}{isual} Spatial Reasoning represents a fundamental cognitive ability in humans~\cite{piaget2013child, moser2015place}. It enables us to derive rich spatial information from observing the world, which is crucial for navigating and interacting with real-world environments. This capacity is not only vital for humans but is also indispensable for advancing key AI domains such as embodied intelligence and autonomous driving. Consequently, achieving human-level visual Spatial Reasoning in machines has been a long-standing pursuit in the research community~\cite{alayrac2022flamingo, dai2023instructblip, liu2025ola, chen2024end, jin2023adapt, bojarski2016end, kendall2019learning, chitta2022transfuser}. However, this objective remains highly challenging, primarily due to the inherent complexity of representing and reasoning about three-dimensional (3D) space.

% Initially, LLaVA~\cite{liu2023visual} enhances the visual understanding capabilities of large language models by using instruction tuning to efficiently align image and text modalities with image-text paired data. Subsequent research~\cite{InternVL3, BAGEL, Qwen3, BLIP, Flamingo, Seed1.5-vl} introduces a staged training paradigm, which incorporates pretraining, continued training, post-training, Supervised Fine-Tuning (SFT), and Reinforcement Learning (RL). The progressive development of these methods significantly improves the performance of VLMs on general VQA tasks.

In recent years, Vision-Language Models (VLMs)~\cite{InternVL3, BAGEL, Qwen3, BLIP, Flamingo, Seed1.5-vl, jin2024llava, li2024llava, chen2025blip3, dai2023instructblip,gpt5,team2023gemini,zeng2025glm} attract significant attention due to the powerful visual understanding and reasoning capabilities, bringing widespread applications in various Visual Question Answering (VQA) tasks~\cite{kuang2025natural, albalak2025bigmathlargescalehighqualitymath, pennino2025reasoning}. Concurrently, an emerging body of work~\cite{SpatialVLM, SpatialRGPT, zhou2025roborefer, Visual-Cot} begins to apply VLMs to Visual Spatial Reasoning (VSR). However, VSR presents a distinct and more formidable set of challenges compared to general-purpose VQA, which often focuses on semantic-level understanding, as illustrated in Fig~\ref{fig:radar}. 
While general VQA targets object recognition or attribute identification, VSR demands intricate reasoning about complex spatial dynamics, such as the relative positions, orientations, distances, and object motion. This need for deep spatial awareness is particularly exacerbated in multi-view and video contexts, which not only amplify the reasoning complexity but also increase the model's propensity for hallucinations. Furthermore, this difficulty extends to data curation. Building robust VSR datasets is substantially more demanding, requiring precise spatial annotations and complex scene analysis that vastly exceed the requirements of traditional VQA benchmarks.

% Visual Spatial Reasoning (VSR) is a critical area of general Visual Question Answering (VQA). These tasks poses unique challenges that distinguish it from traditional Visual Question Answering (VQA). While VQA targets basic semantic understanding, VSR requires models to reason about complex spatial dynamics—the relative positions, orientations, distances, and motion of objects. This need for deep spatial awareness is especially critical in multi-view and video contexts, which amplify reasoning difficulty and increase the frequency of model hallucinations. Accordingly, building VSR datasets is also far more difficult, requiring precise spatial annotations and complex scene analysis that vastly exceed the demands of traditional VQA.
%Unlike traditional VQA tasks, VSR presents unique challenges. VSR not only requires understanding basic semantic information in images, but also demands that models reason about complex spatial relationships and dynamic changes between objects, such as relative positions, orientations, distances, and motion trajectories. These tasks necessitate a deeper level of spatial understanding, especially when dealing with multi-view or dynamic changes in videos, where reasoning becomes more difficult, and hallucination phenomena are more pronounced. Furthermore, constructing VSR datasets is more challenging, as it requires precise spatial relationship annotations and detailed analysis of complex scenes, making it far more demanding than traditional VQA tasks.

Meanwhile, research in VSR advances rapidly, with current efforts concentrating on two primary fronts. On one hand, the community continues to drive innovation in input modalities~\cite{SpatialRGPT}, model architectures~\cite{zhou2025roborefer}, training strategies~\cite{ouyang2025spacer}, and reasoning mechanisms~\cite{Visual-Cot}. %For instance, foundational work like SpatialVLM~\cite{SpatialVLM} constructs image-text pairs that encode spatial relationships using techniques like object detection and depth estimation. Following this, benchmarks such as VSI-Bench~\cite{VSI-Bench} introduce novel task formats, including navigation and temporal-appearance ordering, to assess the performance of VLMs from multiple perspectives. 
On the other hand, there is a dedicated push~\cite{VSI-Bench, Spatial-MM, OmniSpatial, SpatialBench} to develop higher-quality and more diverse datasets that enable comprehensive evaluation of VSR capabilities.% For example, SpatialRGPT~\cite{SpatialRGPT} enhances geometric understanding by incorporating depth data, while Visual-CoT~\cite{Visual-Cot} boosts spatial reasoning by integrating a Chain-of-Thought (CoT) process at inference time.

%Currently, research in VSR is focused on two main directions. On the hand, researchers are dedicated to developing higher-quality and more diverse task-setting datasets to facilitate comprehensive evaluations of VSR tasks. For instance, early work such as SpatialVLM~\cite{SpatialVLM} constructs image-text pairs related to spatial relationships for model training, using techniques such as object detection and depth estimation. Subsequently, VSI-Bench introduces new task settings, such as navigation and temporal-appearance order, evaluating the performance of VLMs in VSR tasks from multiple perspectives. On the other hand, the community continues to explore innovations in model architecture, training strategy, and reasoning mechanisms. For example, SpatialRGPT enhances spatial geometric understanding by incorporating additional depth data, while Visual-CoT~\cite{Visual-Cot} improves spatial reasoning by introducing a Chain-of-Thought (CoT) mechanism during the inference phase.

% The field of VSR has seen rapid development in recent years. 
Despite a wealth of related research, there remains a lack of a systematic review, particularly in terms of methodologies and the specific task settings. 
Crucially, despite the recent proliferation of evaluation benchmarks, the landscape remains highly fragmented. Each existing dataset typically addresses only a narrow and specific set of tasks, thus failing to provide a comprehensive assessment of a model's visual spatial reasoning capabilities.
% Also, the evaluation landscape is highly fragmented, with over \textbf{fifty} open-source benchmarks spanning more than \textbf{thirty} distinct task settings. This fragmented state makes it challenging to conduct unified and objective assessments of VLMs in their spatial reasoning capabilities, complicating research and comparisons.
Given these issues, this paper aims to provide a detailed review of the existing methods and datasets in the field, while also curating, consolidating, and organizing the current benchmarks into a comprehensive VSR evaluation dataset. It provides a convenient, objective, and comprehensive evaluation tool for assessing the spatial understanding capabilities of VLMs.

In summary, this paper makes the following key contributions:
\begin{itemize}
    \item \textbf{Thorough Review of Visual Spatial Reasoning Methods.} We review and analyze existing methods, focusing on input modalities, model architectures, training approaches, and reasoning mechanisms, providing a systematic reference for researchers.
    \item \textbf{A Hierarchical Task Categorization Based on Cognitive Levels.} We introduce a systematic categorization of VSR tasks organized by cognitive levels, outlining their core objectives, inherent challenges, and the current research progress to guide future work.
    %We introduce a systematic categorization of Visual Spatial Reasoning (VSR) tasks organized by their cognitive depth. Our framework divides tasks into three distinct levels: Basic Perception, Spatial Comprehension, and Planning. For each level, we analyze its core objectives, inherent challenges, and the current state of research to provide a clear roadmap for future investigations.
    %\item \textbf{Categorization of Tasks Based on Reasoning Depth:} We organize VSR tasks by their reasoning depth, outlining their objectives, challenges, and current research progress to guide future work.
    \item \textbf{Development of SIBench, a Comprehensive Evaluation Benchmark.} We introduce SIBench, which curates nearly 20 open-source benchmarks covering 23 distinct VSR task settings. Beyond serving as a rigorous and comprehensive evaluation tool for VLMs, our analysis using SIBench reveals that current models exhibit significant deficiencies in VSR tasks. These shortcomings are particularly pronounced in areas such as precise numerical estimation, multi-view reasoning, temporal information processing, and spatial imagination.

\end{itemize}

In the following sections, we provide a comprehensive review of over 150 research papers related to VSR published since 2023. Firstly, we define VSR and outline its research scope in \refsection{sec:background}, followed by a thorough review of existing methodologies in \refsection{sec:methodologies}. Next, we categorize the tasks involved in VSR based on cognitive levels and discuss the primary challenges these tasks present in \refsection{sec:task_settings}. We then introduce a comprehensive evaluation dataset for VSR and present evaluation results for several models in \refsection{our eval}. Finally, we conclude by summarizing future research directions and potential applications in \refsection{sec:discussion}, highlighting key challenges that need to be addressed in the field. 

% We create an open-source repository that provides a taxonomy of all mentioned papers and code links. The repository link is \url{https://github.com/prism-visual-spatial-intelligence/Awesome-Visual-Spatial-Reasoning}.
\section{Background}
\label{sec:background}
\subsection{Research Scope}
In recent years, with the development of Vision-Language Models~\cite{alayrac2022flamingo, yang2023dawn, dai2023instructblip, chen2025blip3, wu2025omnigen2, team2025kwai, liu2025ola} (VLMs) and generative models~\cite{labs2025flux1kontextflowmatching, li2024hunyuan, dhariwal2021diffusion, song2020denoising, ho2020denoising, katzir2023noise}, the realization of spatial intelligence has become increasingly promising. Spatial intelligence manifests in various aspects. First, agents perceive the 3D world through sensor inputs, gaining an understanding of its basic properties, followed by the comprehension of spatial relationships and physical laws. Furthermore, agents can interact with their environment, such as performing spatial navigation and manipulating objects. Additionally, the ability to create and imagine entirely new worlds is also a crucial aspect of spatial intelligence. The scope of spatial intelligence is vast, covering a wide range of tasks and applications, such as spatial reasoning, visual-language localization~\cite{chu2024towards,ju2024video2bev}, embodied artificial intelligence~\cite{zitkovich2023rt, black2024pi_0, kim2024openvla, chi2024universal}, and video world models~\cite{gao2024cardreamer, yang2025driving, zheng2024occworld}.

For agents, image/video input is a readily accessible and cost-effective form of data. For biological organisms, vision is also a vital pathway for spatial modeling. Therefore, this paper focuses on the application of VLMs in spatial reasoning, which involves interpreting spatial information from images, multi-view inputs, or videos, including basic perception, spatial relationship understanding, and planning. It should be noted that tasks involving point clouds~\cite{pan2025omnimanip, ding2024holistic, li2024know, rizvi2024sparc, zhou2025physvlm, yang20253d, qi2024shapellm} as input for multimodal spatial reasoning or pure-text-based~\cite{li2024advancing, mouselinos2024beyond, rizvi2024sparc, wang2024dspy, li2025imagine} spatial reasoning are not within the scope of this research. Additionally, while generative models~\cite{bogdoll2023muvo, cai2023diffdreamer, yang2024worldgpt, bruce2024genie, liu2024world, yin2023nuwa} also reflect spatial intelligence, they differ significantly in their modeling mechanisms and spatial representation form VLMs, and thus are not part of this discussion. Our focus is primarily on the general ability of VLMs in understanding spatial relationships, specific applications like vision-language action models~\cite{liao2025diffusiondrive, li2025generative, jiang2025transdiffuser, tang2025hip} and vision-language navigation models are not the main focus of our study.

\subsection{Related Work}
Several studies have provided comprehensive reviews of the development and evaluation frameworks of Vision-Language Models (VLMs). For example, ~\cite{li2025benchmark} systematically organizes the technological evolution, evaluation benchmarks, and application scenarios of VLMs, highlighting the potential performance fluctuations when handling complex tasks. The study further suggests introducing new evaluation metrics, emphasizing the importance of visual localization and multimodal understanding.

The study~\cite{ma2024llms} investigates the integration of LLMs with 3D spatial understanding, analyzing challenges in data representation, model architecture, and evaluation metrics. It also recommends improvements in these three areas. It should be noted that the study in~\cite{ma2024llms} primarily focuses on using 3D representations as input, rather than image/video input. Zha~\etal~\cite{zha2025enable} similarly explores the spatial reasoning capabilities of LLMs, categorizing tasks based on input modalities such as images, point clouds, and hybrid modalities. Additionally, some studies~\cite{qi2025beyond, chen2025spatial, zhang2024vision} have revealed shortcomings in the spatial reasoning abilities of VLMs, citing issues such as the lack of effective spatial attention mechanisms or the limited ability of existing attention mechanisms to align with object positions. Sapkot~\etal~\cite{sapkota2025vision} reviews the development of Vision-Language Action Models (VLA) and points out challenges in real-time control, dataset bias, and system integration when applying these models in practice. Researchers in~\cite{xie2025vlms} discusses the application of VLMs in autonomous driving scenarios, noting that the reliability of VLMs remains insufficient in complex traffic environments. Guo~\etal~\cite{guo2024survey} summarizes the technological evolution and applications of VLMs, highlighting challenges in data quality and complex tasks within the biomedical field.

% The scope of this study is on the spatial reasoning abilities of VLMs, specifically targeting pure visual capabilities, where the inputs are single images, multi-view images, or videos. Our aim is to review current advancements in visual spatial reasoning and categorize spatial reasoning into three cognitive levels: basic perception, spatial understanding, and planning. We provide a detailed and systematic analysis of existing task formulations. Additionally, we have gathered and curated a high-quality, comprehensive dataset from open-source benchmarks to evaluate the spatial reasoning abilities of VLMs.

\begin{figure*}[!t]
\centering
\includegraphics[width=7in]{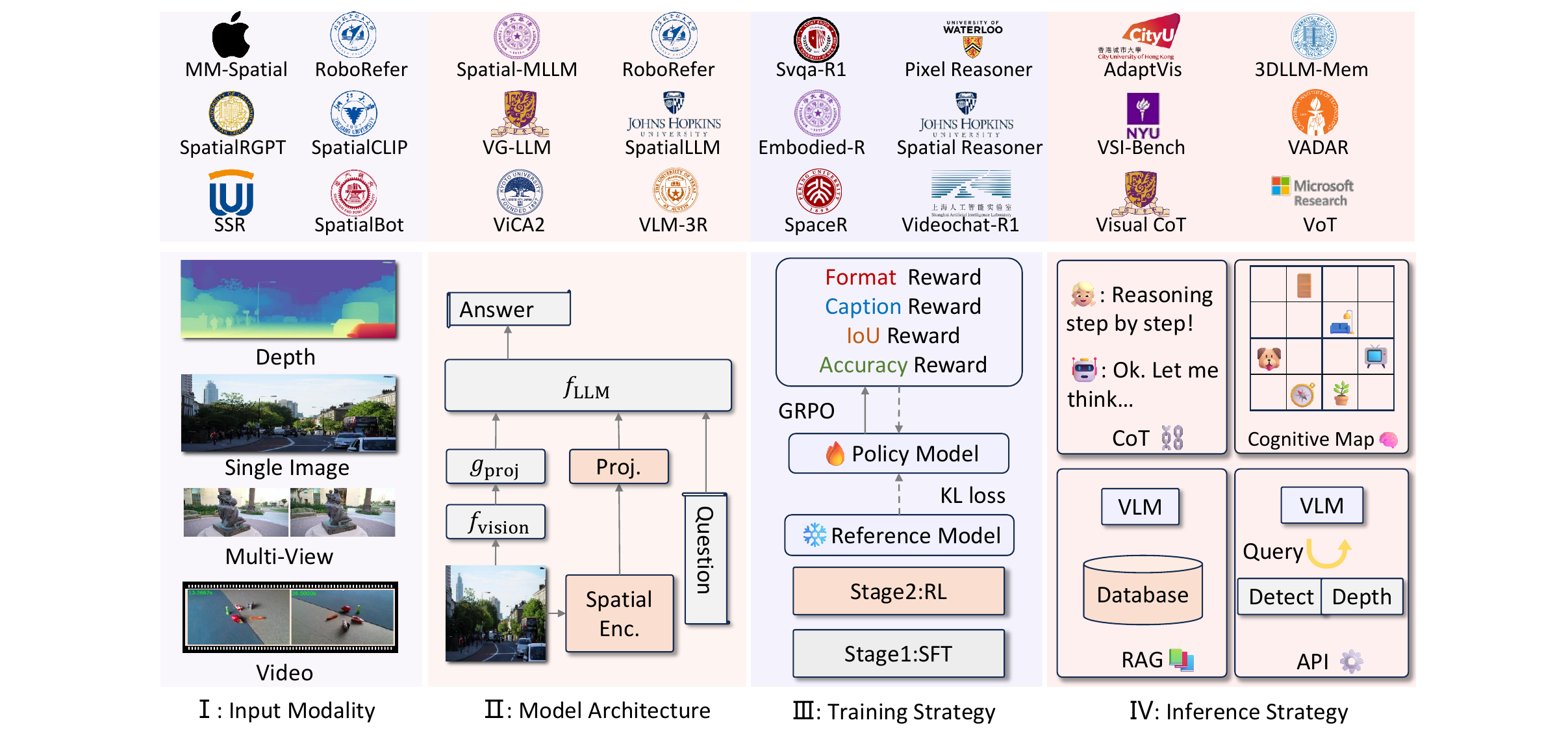}
\caption{
\textbf{An overview of the primary methods} (Bottom).
I: Incorporating an additional input modality, such as depth maps. 
II: An additional spatial encoder is incorporated into the model architecture to provide 3D information.
III: Leveraging Reinforcement Learning to improve generalization.
IV: The inference phase employs methods such as cognitive maps to perform structured reasoning.
\textbf{Representative methods for the four categories} (Upper).
} 
% 字体放大
\label{fig:methods}
\end{figure*}

\section{Methodologies}
\label{sec:methodologies}
% \subsection{Overview}
VSR requires that VLMs not only understand semantic information and localize targets but also reason about the spatial relationships between multiple objects, imagine 3D spatial structures from 2D images, and have the potential for dynamic prediction. In this chapter, we analyze existing research methods and summarize them into four areas for improvement: input modalities, model architecture, training strategy, and inference methods (see \reffig{fig:methods} for more details).

\subsection{Naive Solution}
% image-only
To better understand VSR methods, we begin by introducing the commonly used naive solutions. Similar to general VLMs, a VSR model takes an image $I$ and a spatially-related question $Q$ as input to produce an answer $A$. The overall process can be conceptualized as modeling the conditional probability $P(A | I, Q)$. The architecture typically consists of three core components: a vision encoder $f_{\mathrm{vision}}$, a projection module $g_{\mathrm{proj}}$, and a Large Language Model (LLM) $f_{\mathrm{LLM}}$.

Generally, the vision encoder processes the image to extract a set of feature vectors $F_v = f_{\mathrm{vision}}(I)$, where $F_v \in \mathbb{R}^{N \times D_v}$ is a sequence of N feature vectors, each with dimension $D_v$, capturing both semantic and spatial information. Concurrently, the text question $Q$ is converted into embeddings $E_q \in \mathbb{R}^{M \times D_t}$ by text encoder. Given the potential mismatch in dimensions and feature spaces between the two modalities, the projection module is employed to align the visual features with the text embedding space. This module transforms $F_v$ into a new representation $F'_v = g_{\mathrm{proj}}(F_v)$. This projection is usually implemented with a few linear layers or a cross-attention mechanism, mapping the visual features so that $F'_v \in \mathbb{R}^{N \times D_t}$, aligning their dimension with the text embeddings. Finally, the features from both modalities are concatenated and fed into the LLM, which then autoregressively generates the answer: $A = f_{\mathrm{LLM}}(\text{concat}(F'_v, E_q))$. Notable examples of this structure include SpatialVLM\cite{chen2024spatialvlm}, LLaVA-VSD\cite{jin2024llava}, and other related works\cite{liu2025coarse, wen2024can, tang2024sparkle, liu2022things,ogezi2025spare, wang2024root, yuan2024robopoint, liu2025reasongrounder, team2025robobrain, xu2025multi, xu2025llava, li2024proximity, zhao2024can}. This naive solution is both concise and efficient. Subsequent work, focusing on the specific characteristics of visual-spatial reasoning tasks, typically introduces improvements in four key aspects: input modalities, model architecture, training strategies, and inference methods.

% \subsection{Depth-Augmented Spatial Reasoning Models}

\subsection{Assisted Input Modalities}
RGB images are a planar projection of the 3D world, providing continuous texture information. However, during the projection process, real 3D structural information is lost, which is one of the challenges in visual spatial reasoning tasks~\cite{wang2025spatialclip}. 3D point clouds offer a better representation of structure, but their data scale, diversity, and quality fall far behind that of images. As a result, some approaches~\cite{zhou2025roborefer, SpatialRGPT} attempt to strike a balance between 3D and 2D. These methods, often referred to as 2.5D~\cite{jin2023adapt}, utilize depth maps as an additional modality. The motivation is that combining depth maps with images can yield a representation similar to point clouds, enabling the recovery of 3D structure in the real world, while the input images retain complete complex texture. Furthermore, such RGB-D data is relatively easy to obtain, especially in embodied scenarios.

Due to the substantial distribution gap between depth maps and natural images, effectively encoding depth maps is by no means a trivial task. Additionally, the visual encoders of most VLMs are only trained on text-image pairs, and simply concatenating RGB and depth features may negatively affect performance~\cite{SpatialRGPT}.

To address these issues, SpatialRGPT\cite{SpatialRGPT} first replicates the single-channel depth map $D$ to match the three channels of an RGB image. It then employs a shared vision encoder, $f_{\mathrm{vision}}$ to process both modalities, yielding features $F_v = f_{\mathrm{vision}}(I)$ and $F_d = f_{\mathrm{vision}}(D)$. These features are subsequently passed to different projectors, a design which ensures the model can still function normally without depth input and avoids the need for large-scale image-depth data training. Building on this, RoboRefer~\cite{zhou2025roborefer} uses a completely separate Depth encoder $f_{\mathrm{depth}}$ to handle depth inputs, thereby avoiding modality interference. This architecture choice can be expressed as: $F_v = f_{\mathrm{vision}}(I)$ and $F_d = f_{\mathrm{depth}}(D)$. This separation helps preserve general VQA performance while enhancing the perception of depth cues. SSR~\cite{liu2025ssr} also aims to integrate depth information in a similar manner.
%To address these issues, SpatialRGPT\cite{SpatialRGPT} first replicates the depth map three times along the channel dimension, and then shares a single encoder between RGB and depth, with the resulting RGB and depth features being processed by different connectors. This design ensures that the model can still function normally without depth input and avoids the need for large-scale image-depth data training. Building on this, RoboRefer\cite{zhou2025roborefer} uses a separate Depth Encoder to handle depth inputs, thereby avoiding modality interference and preserving general VQA performance while enhancing the perception of depth cues such as distance, near-far relationships, and perspective-based size variations. Similar approaches to this method include SSR\cite{liu2025ssr}, which also aims to integrate depth information in a similar manner. 
In addition to using modular designs for feature fusion, SpatialCLIP\cite{wang2025spatialclip} is inspired by point cloud reconstruction and 3D point cloud architectures. It utilizes depth information to lift 2D tokens to 3D space, thereby better modeling the 3D spatial relationships. Specifically, for each transformer layer, it first lifts the 2D tokens to 3D voxels based on the average depth of each patch, and then introduces 3D depth convolutions to capture spatial patterns in the 3D space, thus providing enhanced spatial information.% 这里需要加公式化描述

In addition to processing at feature space level, some models introduce an explicit, symbolic reasoning step by interacting with external tools. SpatialBot~\cite{cai2024spatialbot} provides a method for querying depth values from a given depth map $D$ through an API. The process begins when the model receives an image $I$ and a question $Q$. Based on the question, SpatialBot first determines if depth information is required. If so, it identifies the target coordinates $(x, y)$ in the image relevant to the question. For instance, when asked about an object's depth, it first computes the objects's bounding box, $B_{\mathrm{obj}}$, and then determines its center point: $(x_c, y_c) = \text{Center}(B_{\mathrm{obj}})$. Subsequently, the model generates a symbolic API call, such as $\text{Depth}(x_c, y_c)$. This triggers an external function that queries the depth map $D$ to retrieve the corresponding depth value $d$: $d = \text{API\_Query}(D, (x_c, y_c))$. This numerical value $d$ is then converted back into a natural language string, $T_d$. Finally, SpatialBot generates the final answer A by processing a new augmented prompt that concatenates the original question with the retrieved depth information. This method of querying an API and re-inputting the information as text is further explored by MM-Spatial~\cite{daxberger2025mm}. Their experiments also revealed that this ``Query-and-Re-input" cycle yields significantly better results for spatial reasoning tasks in VLMs. While this class of methods often achieves high accuracy, its pipeline is also quite complicated.

% \subsection{Dual Visual Encoder}
\subsection{Model Architecture Optimization}
Numerous existing Vision-Language Models, such as LLaVA\cite{liu2023visual} and its subsequent versions~\cite{li2024llava, xu2025llava, guo2024llava} employ contrastive learning for pretraining visual encoders, generating compact and expressive visual embeddings. These models align well with natural language, demonstrating strong performance in tasks like image captioning and general visual question answering. However, similar to the contrastive learning used in CLIP\cite{radford2021learning}, they primarily optimize for global semantic alignment and often neglect fine-grained spatial reasoning. Consequently, the visual embeddings produced by these encoders, while capturing an image's overall semantic gist, provide only a coarse representation of its contents and are inherently limited in encoding precise spatial information.

% \begin{figure*}[!t]
% \centering
% \includegraphics[width=7in]{figure/methods_table.png}
% \caption{Timeline of Representative VSR Benchmarks. Purple indicates novel evaluation methods and green denotes methodological improvements.}
% \label{fig:methods table}
% \end{figure*}

To overcome this limitation, a powerful strategy has emerged: the dual-visual-encoder architecture. This design addresses the issue by creating two complementary visual pathways. While the original encoder provides the high-level semantic context, a second, specialized encoder is introduced to supply the fine-grained details. For single-image inputs, models like SpatialLLM\cite{ma2025spatialllm} and ViCA2\cite{feng2025towards} incorporate additional encoders pre-trained on detail-oriented tasks. The reason for choosing models like MAE\cite{he2022masked}, DINO v2\cite{oquab2023dinov2}, and SAM\cite{kirillov2023segment} is that their training objectives, such as masked image reconstruction or large-scale segmentation, force them to learn rich, pixel-level features that the primary semantic encoder ignores. For inputs with inherent 3D information, like video or multi-view images, this approach is taken a step further. Standard fine-grained encoders may not be sufficient to interpret the geometric cues embedded in motion and parallax. Therefore, several studies use 3D reconstruction models as the second encoder. For instance, VLM-3R~\cite{VSTiBench} incorporates CUT3R~\cite{wang2025continuous} to model the scene's underlying 3D geometry from multiple views. By doing so, the model gains access to much richer representation that include depth and structure. Similarly, VG-LLM~\cite{zheng2025learning} and Spatial-MLLM\cite{wu2025spatial} use advanced reconstruction networks like VGGT\cite{wang2025vggt} to provide the language model with an even more sophisticated understanding of 3D space, significantly enhancing its spatial reasoning capabilities.

% 需要详细介绍一下connector
% 公式化阐述一下，增加细节
% EgoDTM 是对visual encoder增加一个深度的监督信息

% \subsection{Reinforcement Learning based Methods}
\subsection{Training Strategy Optimization}
VSR goes beyond perceiving static structures, typically requiring dynamic, multi-step reasoning grounded in commonsense knowledge.
This multi-step nature makes it a natural fit for reinforcement learning, especially Group Relative Policy Optimization (GRPO)~\cite{guo2025deepseek}, which has already shown success in enhancing the textual reasoning and generation capabilities of LLMs and VLMs.
Inspired by this success, a growing body of work \cite{li2025videochat,wang2025svqa,zhao2025embodied,ouyang2025spacer} is now investigating RL's potential for spatial reasoning tasks. 
However, unlike mathematical reasoning~\cite{albalak2025bigmathlargescalehighqualitymath} or code generation~\cite{pennino2025reasoning} tasks, where correctness can be directly verified, spatial reasoning lacks a clear, well-defined reward signal. This makes it difficult for RL to acquire spatial reasoning abilities effectively, leaving the application of RL to VSR a challenging and open problem.

To deal with such challenges, current works generally adopt two strategies.
The first category~\cite{su2025pixel,zhou2025roborefer,zhao2025embodied,ma2025spatialreasoner} adopts a two-stage paradigm, in which a Supervised Fine-Tuning (SFT) stage is introduced before Reinforcement Learning Fine-Tuning (RLFT) to provide explicit supervision for spatial reasoning.
The SFT stage serves as a warm start for training, while the RLFT stage empowers models with strong generalization ability.
For example, RoboRefer~\cite{su2025pixel} first trains VLMs with depth, spatial understanding data, and instruction tuning at the SFT stage and further fine-tunes VLMs with multi-step reasoning data at the RLFT stage.
The second category adopts a one-stage paradigm, which focuses on designing task-specific reward functions tailored for spatial reasoning in RL training.
Considering the invariance and variations between original and flipped images, Wang~\etal~\cite{wang2025svqa} extend GRPO to Spatial-GRPO, comparing rewards between the original and flipped groups and penalizing the group with a higher score.  
SpaceR~\cite{ouyang2025spacer} explicitly constructs an object-centric map and defines a reward function based on this map to establish quantitative feedback for spatial understanding.
In the video spatial reasoning domain, Li~\etal~\cite{li2025videochat} incorporate additional explicit IoU-based and recall-based rewards into GRPO.
By training with different combinations of reward functions in GRPO, these methods achieve improved performance and generalization ability compared with SFT-trained variants.

% 分类
% - 设计更好的RL: 
%   - VideoChat-R1: motivation: rule-based reward systems and multimodal temporal dependencies-> Spatio-Temporal reasoning: IoU reward; Recall reward 
%   - SVQA-R1: motivation: extend the R1 framework to spatial reasoning-> Mirror-Consistent Reasoning: Semantic-aware Reward 

%   - SpaceR: motivation:  explicit reward signals for spatial information map reward
% - 用RL增强小模块：(略，简单一句)
%  - Pixel Reasoner
%   - Embodied-R: motivation: trade-off between reasoning speed and model size -> small-scale language model reasoning + Logical Consistency Reward (Logical Consistency between reference model and small LM)
%  - SpatialReasoner
%  - RoboRefer

% \subsection{Others} %round off cot
\subsection{Inference Strategy Enhancements}
VSR requires VLMs to conduct a series of steps, such as scene understanding, object grounding, and various forms of reasoning, including relationship and trajectory. Therefore, introducing intermediate procedures (\eg, building a map for the scene and integrating multimodal features) can effectively strengthen both visual and spatial capabilities. To provide a clear taxonomy, we categorize existing methods according to the source of their enhancement signals.

\noindent\textbf{Internal enhancement.}
Internal enhancement refers to leveraging pretrained world knowledge inherent in LLMs without injecting additional external knowledge. According to the objects of enhancement, we divide this into three categories.
(1)~\emph{Multimodal Chain-of-Thought (CoT).}
Prompting LLMs to conduct several steps of linguistic reasoning has shown effectiveness~\cite{huang2022language,suris2023vipergpt}. 
Inspired by this, multimodal CoT has been introduced for VSR tasks, enabling joint reasoning over visual and textual spaces and thereby enhancing spatial awareness during intermediate reasoning. To elicit and enhance the reasoning steps, previous works~\cite{wu2024mind,li2025imagine} propose to visualize the intermediate steps of reasoning.
For instance, Wu~\etal~\cite{wu2024mind} prompts the pretrained VLMs to generate reasoning traces $z_{1....i}$ and spatial visualizations $v_{1....i}$ in a separate manner as follows:

\begin{equation}
    z_i \sim p_{\theta}(v_i|x,z_{1....i-1},v_{1....i-1}),
    \label{eq:sep_z}
\end{equation}
\begin{equation}
    v_i \sim p_{\theta}(v_i|x,z_{1....i},v_{1....i-1}).
    \label{eq:sep_v}
\end{equation}
% 这个是需要训练的
Inspired by ReAct~\cite{yao2023react}, Yao~\etal~\cite{li2025imagine} propose to generate intermediate spatial visualizations and reasoning traces simultaneously as follows:
% , shown in~\ref{eq:unified_vz}.
\begin{equation}
    v_i, z_i \sim p_{\theta}(v_i, z_i|x,z_{1....i-1},v_{1....i-1}).
    \label{eq:unified_vz}
\end{equation}
Both methods aim to make reasoning steps interpretable.
To deal with the lack of intermediate supervision for spatial CoT,
Visual CoT~\cite{shao2024visual} collects a dataset consisting of detailed spatial reasoning steps and bounding boxes of Regions of Interest (RoI), serving as ground-truth for CoT supervision.
% collects a dataset consisting of GT with detailed spatial reasoning steps and bounding boxes of Regions of Interest (RoI). 
Based on the dataset, the VLMs can be trained to extract visual tokens from the RoIs and enhance reasoning steps based on both RoIs and the entire input, following the ground-truth chain of thought. 
This design enables the model to dynamically focus on the relevant regions during spatial reasoning steps.
% In this way, this method can dynamically focus on and zoom in on the relevant regions during the reasoning process.
(2)~\emph{Scene Representation.}
Inspired by {the hippocampus's central role in contextual memory}~\cite{nadel2008hippocampus}, some methods~\cite{VSI-Bench, MINDCUBE} propose to generate internal representations of spaces and conduct reasoning based on these.
% cog map
Given a video input and a question, Yang~\etal~\cite{VSI-Bench} first construct a cognitive map consisting of object center positions in a grid format. Then the predicted map is leveraged to answer the question, achieving robust results regarding local distance awareness.
% Then they use the predicted map to answer the question, receiving robust results of local distance awareness.
% 具身
Similarly, when exploring scenes in embodied tasks, a powerful internal representation is also useful to maintain past and current observations.
% 3d voxel -> 2d cog map
For example, some works~\cite{ren2024explore,hu20253dllm} adopt 3D voxels as internal scene representation, on which the occupancy and exploration status are updated first. Afterwards, voxels are projected onto a 2D semantic map~\cite{ren2024explore}. For the unexplored areas, this map is utilized to predict probabilities over three candidate directions and to estimate whether further exploration is worthwhile based on current observations.
(3) \emph{Attention re-distribution.} 
During reasoning, there exists an imbalance between visual and textual attention.
% motivation
Specifically, image tokens constitute 90\% of the input sequence but only account for 10\% of the total attention, leading to {misalignment} between the {actual spatial layout of objects in images} and {instruction prior in texts}~\cite{chen2025spatial}.
% actual geometric distribution of objects~\cite{chen2025spatial}.
% method
{To deal with this imbalance}, AdaptVis~\cite{chen2025spatial} introduces a dynamic attention redistribution strategy {that leverages output logits as guidance}. 
% enhancing in an internal way.
% Specifically, AdaptVis uses the logits of the generated output as guidance. 
When the guidance is low, the attention distribution is smoothed, encouraging the model to explore a broader range of inputs. {In contrast}, when the guidance {signal} is high, the attention distribution is sharpened, {directing the model to focus on key objects}. 
% focus the model on key objects.

\noindent\textbf{External enhancement.}
% \noindent\textbf{Training-free Methods.}
External enhancement supplements VLMs with knowledge beyond their pretrained parameters to understand illogical or uncommon spatial relationships.
{One representative direction is the use of \emph{multi-agent system}.}
{Marsili~\etal\cite{marsili2025visual} designs a cooperative system where different agents interact in a code policy manner: API generation agents first decompose the query into sub-questions, while program synthesis agents receive these sub-questions and generate code to solve each problem.}
% motivation
{Another line of work is \emph{Retrieval Augmented Generation (RAG)}}.
{For example, Yu~\etal~\cite{yu2024rag} retrieves templates for subjects-objects relationships as well as other {spatial descriptions}, and then integrates these templates into the inference stage via in-context learning.}

% 其他中挑重要的
% 构造map丢到这里
\section{Task Settings}
\label{sec:task_settings}

\begin{figure}[!t]
\centering
\includegraphics[width=3.5in]{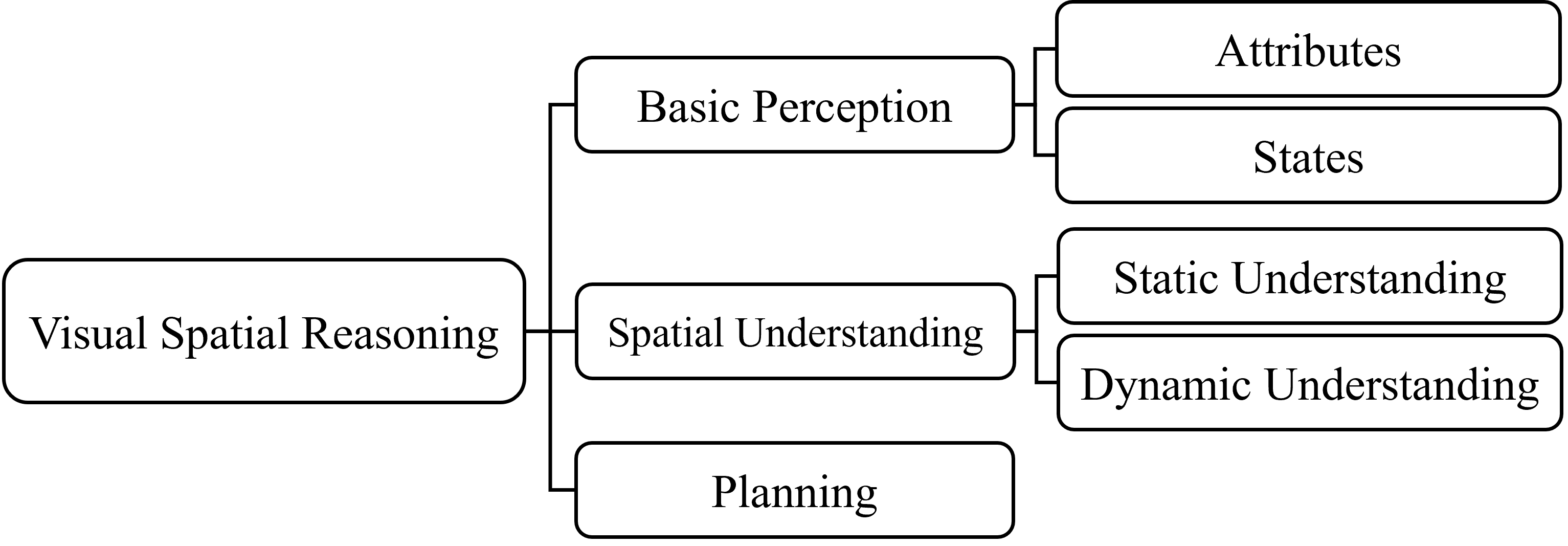}
\caption{Taxonomy of visual spatial reasoning according to cognitive levels.}
\label{fig:reasoning_levels}
\vspace{-0.2in}
\end{figure}

\begin{figure*}[!t]
\centering
\includegraphics[width=7in]{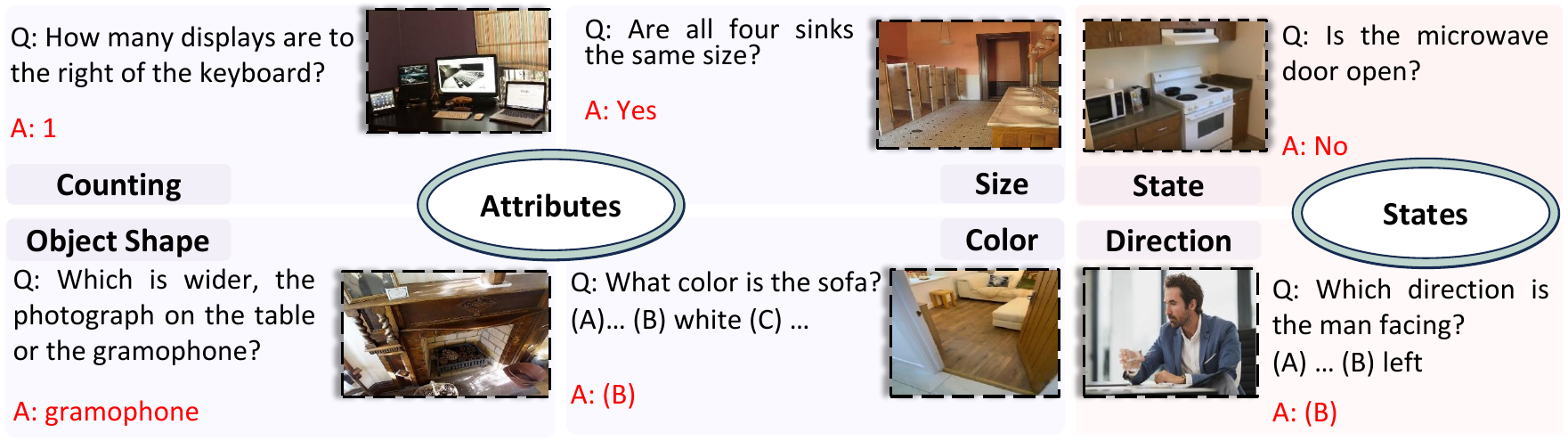}
\caption{\textbf{Task Settings for Basic Perception}. Basic perception tasks are categorized into static and state attributes based on whether the attribute is subject to change.}
\label{fig:basic_perception_settings}
\end{figure*}

While the systematic evaluation of VSR capabilities in VLMs has gained significant attention, current evaluation baselines face several challenges. First, different benchmarks classify task settings at varying levels of granularity, lacking a systematic framework. Second, task settings are scattered across various benchmarks, making it difficult to assess VSR capabilities comprehensively. 
% Third, the quality of test data is inconsistent, leading to unreliable evaluations.

In this section, we first provide a comprehensive overview of VSR task settings, organizing it into three cognitive levels: Basic Perception~\refsection{sec:perception}, Spatial Relation Reasoning~\refsection{sec:understanding}, and Spatial Planning~\refsection{sec:planning}, as shown in \reffig{fig:reasoning_levels}. The classification is based on levels of reasoning, where basic perception concerns the attributes or states of a single object or category of objects, spatial understanding involves the spatial relationships among multiple objects, and planning refers to seeking satisfactory solutions under spatial constraints. For each level, we introduce the definition of the corresponding task and the challenges it presents. 
% In \refsection{our eval}, we curate and integrate high-quality data from existing benchmarks to create the most comprehensive evaluation baseline to date. We then conduct thorough evaluations of both state-of-the-art VLMs, providing insights into their performance.

\subsection{Basic Perception}
\label{sec:perception}
Current Vision-Language Models (VLMs) are typically first pre-trained on vast corpora of text data, where fundamental object concepts from world knowledge are represented and associated within an abstract vocabulary. Subsequently, through multimodal training, connections are established between this abstract vocabulary and continuous visual representations or discretized image tokens. This process endows them with basic visual question answering (VQA) capabilities.
However, compared to the goal of achieving spatial intelligence, which entails a fine-grained understanding of the 3D world, current VLMs still exhibit deficiencies in foundational object perception. These shortcomings manifest specifically in the perception of object attributes and states. (See \reffig{fig:basic_perception_settings} for an overview of basic perception tasks.)
\subsubsection{Attributes}
Attributes of objects refer to characteristics that remain relatively stable and are unlikely to change in the short term. These typically include shape, color, size, and quantity. The shape of an object is essential for distinguishing between different objects and understanding scenes. Object shape recognition tasks typically begin with simple 2D geometric figures, which include two types of task setups: whole-shape recognition~\cite{GeoMeter,Shapeworld} and sub-shape recognition~\cite{VisOnlyQA}. MM-Bench~\cite{MM-Bench} increases the difficulty by requiring models to recognize object shapes in real images. Additionally, some datasets use multi-view images as input, such as MMSI-Bench~\cite{MMSI-Bench}, which asks the model to deduce an object's shape from multiple views, synthesizing the information to infer the shape from a fixed viewpoint. Color also plays a critical role in object recognition, and color-related tasks are generally simpler, requiring the model to identify the color of a specific object~\cite{MM-Bench, Spatial457, OpenEQA}. In contrast, perceiving the size of an object is more complex. Some studies present two objects for comparison, asking the model to determine which is taller, shorter, wider, narrower, larger, or smaller. Other research requires the model to estimate an object's size quantitatively, which involves using commonsense knowledge for direct measurement~\cite{VSI-Bench, SpatialRGPT, Open3DVQA}. In OMNI3D-Bench~\cite{Omni3D-Bench}, a reference object's physical measurement is provided as a standard. Object quantity recognition~\cite{PhysBench, GeoMeter, Spatial457, SPHERE, SITE, OST-Bench, OSR-Bench, SpatialBench} is one of the more challenging tasks in static attribute understanding. It depends on distinguishing between objects based on attributes such as shape, color, and size. This makes hallucination phenomena more pronounced in tasks involving object quantity recognition.

\subsubsection{States}
In contrast to attributes, states refer to characteristics of objects that are subject to change. Common state attributes include an object's posture, orientation, and its open/closed status. In OpenEQA~\cite{OpenEQA}, the task requires the model to estimate the open/closed state of a microwave door, while other studies focus on determining an object's front-facing orientation in space~\cite{Spatial-MM, Spatial457, Super-CLEVR-3D}. Additionally, some research~\cite{LEGO-Puzzles, OST-Bench} introduces a temporal dimension, requiring the model to understand how states evolve over time. For instance, this might involve describing changes in an object's rotational angle from the past to the present, or transformations in its shape resulting from specific actions~\cite{EOC-Bench}. While the perception of state attributes typically involves describing the properties of a single object or category of objects, and thus falls under basic perception, it is more complex than understanding static attributes. This complexity arises from the diversity of state attributes and the need for commonsense reasoning.

\begin{figure*}[t]
\centering
\includegraphics[width=7in]{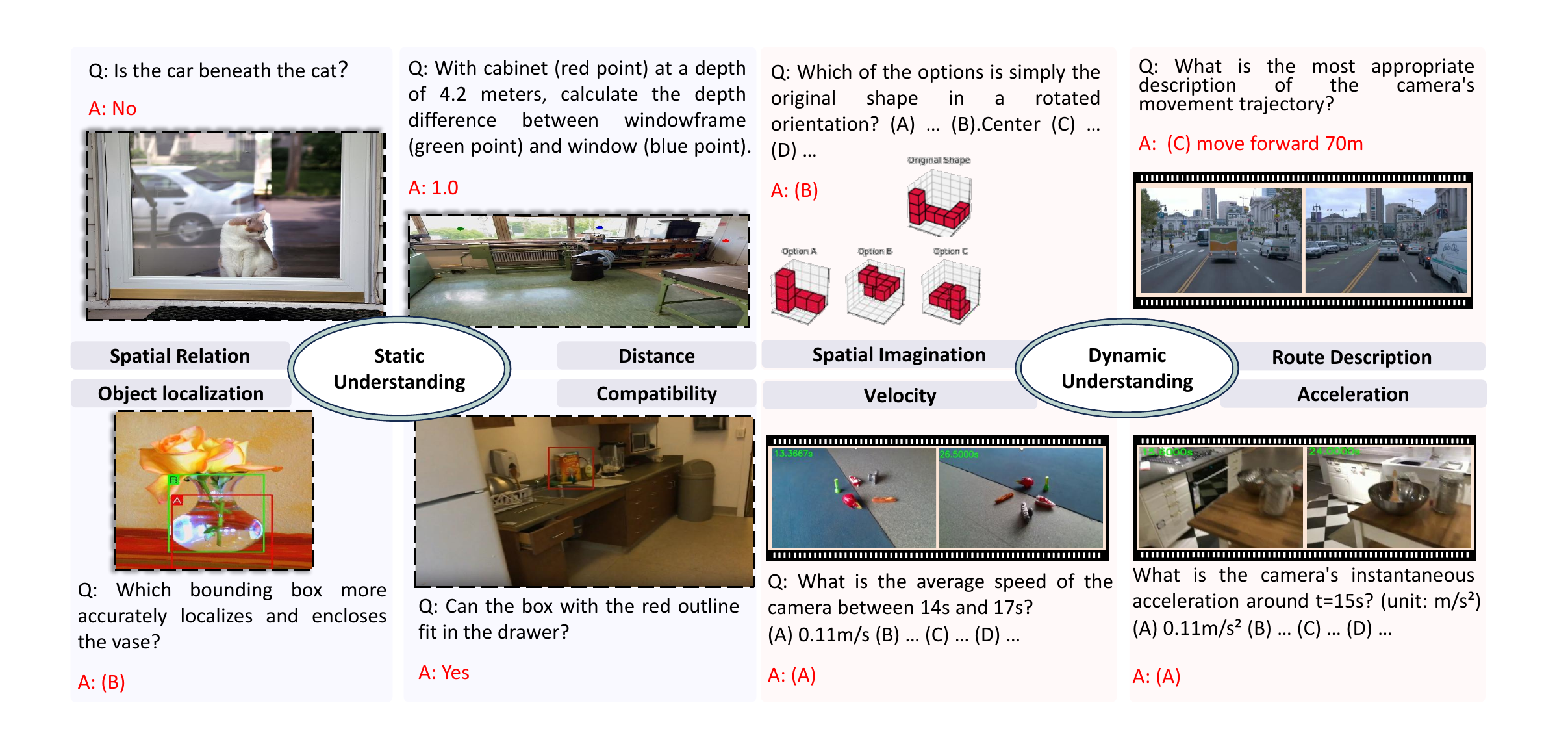}
\caption{\textbf{Categorization of Spatial Understanding Tasks}. Spatial understanding tasks are divided into static and dynamic understanding. Dynamic understanding tasks are characterized by viewpoint shifts or a temporal component.}
\label{fig:spatial_understanding_settings}
\end{figure*}
\subsection{Spatial Understanding}
\label{sec:understanding}
Compared to basic perception, spatial understanding requires the model not only to passively perceive the properties of individual objects or object categories, but also to comprehend the relationships between multiple objects. Furthermore, depending on whether the reasoning object includes a temporal dimension or viewpoint changes, we classify spatial understanding into static understanding and dynamic understanding. See \reffig{fig:spatial_understanding_settings} for an overview of spatial understanding tasks.
\subsubsection{Static Understanding}
The input to static understanding tasks consists of a single image and the corresponding question, with no temporal or viewpoint changes. The tasks primarily include understanding the spatial relationships between objects, object localization, distance measuring, and spatial compatibility reasoning.

\noindent\textbf{Spatial Relationships.}
Early works in the paper VSR\cite{VSR} identified 66 types of spatial relationships \emph{e.g., under, in front of, facing}. The test data consists of image-text pairs, where the text includes a description and a label indicating correctness, such as: \textless Caption: The cow is ahead of the person. Label: False. \textgreater. This requires VLMs to first localize the two objects in the image and then determine whether the abstract relation `ahead of' holds. Later studies introduced additional positional relationships, such as occlusion, reach, and containment. The difficulty in these tasks often lies not in object recognition but in understanding specific spatial relationships. The model must establish connections between the spatial structural relations between objects and the corresponding relation terms.

\noindent\textbf{Object Localization.}
The object localization task requires the model to accurately identify the position of a specified target within an image. The specific localization methods include detecting the precise spatial coordinates of the target object, determining which bounding box best fits the target, or estimating the coordinates of key points on the target. The BLINK~\cite{BLINK} test sample provides an image along with a labeled bounding box for the target object, where the model chooses the bounding box that best aligns with the target from multiple options. The SITE~\cite{SITE} involves the model locating a specific object based on a description or hint and estimating the coordinates of its 2D bounding box. Furthermore, the STI-Bench~\cite{STI-Bench} requires the model to localize the 3D bounding box of the target object within a video. Object localization tasks are also included in other benchmarks~\cite{OpenEQA, MM-Bench, SpatialBench, Visual-Cot, OpenEQA}. The challenge in these tasks lies in the model's need to possess strong object detection capabilities while also handling various complexities in the background and interference, such as partial occlusion and similar objects.

\noindent\textbf{Distance Measuring.} 
Distance measuring tasks aim to estimate the distance between objects in an image or the distance from an object to the camera, which is also referred to as depth estimation. Distance measuring typically involves two forms: qualitative judgment of proximity and quantitative numerical estimation. For instance, in the paper BLINK~\cite{BLINK}, the task requires the model to determine which of two specified points is closer to the camera; in SPAR-Bench~\cite{SPAR-Bench}, the task asks the model to judge which object is closer to or farther from a reference point (such as another object or an observer) based on the 3D center point position of the objects. In addition to qualitative distance analysis, SpatialVLM~\cite{SpatialVLM} employs an automated data construction pipeline to obtain distance data, enabling quantitative distance estimation. In the paper SpatialBench~\cite{SpatialBench}, desktop-level distance estimation is introduced for use in embodied scenarios. Subsequently, many studies have incorporated distance measurement as a crucial task in spatial reasoning \cite{Q-Spatial, SpatialRGPT, All-Angles-Bench, SPACE, VSTiBench, STI-Bench, VSI-Bench, RoomSpace, SPHERE, Proximity-110K, Omni3D-Bench, Open3DVQA, VSI-100K, SpatialScore, PhysBench, SITE}. The challenge of this task lies in the need for the model to infer spatial relationships using implicit cues such as structural information in the scene, object size ratios, and physical knowledge. Furthermore, varying camera parameters may introduce misguidance in distance perception.

\noindent\textbf{Spatial Compatibility Reasoning.} 
Spatial compatibility reasoning tasks require VLMs to determine whether one object can fit the spatial properties or another after thoroughly understanding the object's shape, size, and spatial relationships. In real-world scenarios, this includes determining whether an object can fit into a given space~\cite{OpenEQA} (\emph{'Can another cookie jar fit on the cookie jar shelf?'}). This task assesses the model's reasoning abilities concerning spatial constraints and physical adaptation relationships. Additionally, OmniSpatial~\cite{OmniSpatial} introduces reasoning about motion processes, such as determining whether an object can pass through a narrow passage. Some synthetic datasets also require the model to understand layout constraints and graphical combination rules in a 2D space, enabling it to provide fitting solutions. The difficulty of this task lies in the model's need to not only perceive the size and shape of objects but also reason about physical world common sense, such as ``\textit{an object too large cannot fit into a small space}'' or ``\textit{some uniquely shaped objects cannot be stacked together}''.

\subsubsection{Dynamic Understanding.}
Dynamic understanding typically involves inputs such as multi-view images or videos, introducing the time dimension or changes in viewpoint. Compared to static understanding, dynamic understanding is more complex and variable. The primary tasks include trajectory description, velocity and acceleration estimation, as well as spatial imagination.

\noindent\textbf{Trajectory Description.}
The goal of the trajectory description task is to identify and describe the dynamic changes in the object or camera pose over a time sequence based on visual inputs, generating structured language descriptions. In this context, the trajectory refers to the continuous recording of the camera's pose over time. The descriptions can be divided into two categories based on the observer (camera viewpoint) and the observed world (objective world): one describes the camera's orientation and pose, and the other describes the objects within the observed world. In STI-Bench~\cite{STI-Bench}, the model is required to infer the position and orientation of the camera at a specific time based on video input and initial pose information. In the Ego-ST Bench~\cite{Ego-ST}, the test requires the model to qualitatively describe changes in direction, such as ``\textit{turn left first and then turn right}''. Additionally, multi-view reasoning tasks~\cite{MMSI-Bench, OmniSpatial, SpatialScore, BLINK, STI-Bench, VSTiBench, CameraBench, All-Angles-Bench, SPARE3D} involve numerous estimations of quantified camera parameters. Another category of trajectory description pertains to the observed world, involving tasks like recognizing changes in the direction of other objects in the video, such as ``\textit{the car changes from moving straight to turning left}''. In the VSI-Bench~\cite{VSI-Bench}, a task is designed that requires the model to determine the order of appearance of objects in a spatiotemporal sequence. Similarly, in the SPACE~\cite{SPACE}, multiple landmarks are introduced, and the model is required to determine their sequence. The challenges in these tasks arise from two key issues: first, the model must infer 3D spatial from 2D RGB inputs, which involves geometric principles such as perspective projection, a problem that cannot be resolved by simple image feature matching alone. Second, the estimation errors accumulate as the video sequence length increases, requiring the model to have dynamic error correction capabilities. This is a significant challenge in video pose estimation.

\noindent\textbf{Velocity and Acceleration Estimation.}
In dynamic spatial reasoning tasks, models are required to estimate motion parameters, such as velocity and acceleration, which are crucial for subsequent planning and decision-making. The OmniSpatial~\cite{OmniSpatial} requires models to qualitatively assess the speed and acceleration of objects, like ``\textit{Which car, A or B, is moving faster?}'' or ``\textit{Is the acceleration of the ball on the slope increasing?}''. In contrast, STI-Bench directly asks models to estimate the numerical values of velocity and acceleration from video, such as ``\textit{What is the average speed of the camera?}'' or ``\textit{How quickly is the ball accelerating?}''. Additionally, STI-Bench~\cite{STI-Bench} categorizes tests into three different environments: desktop, indoor, and outdoor, based on varying accuracy requirements for motion parameter estimations.
In DynSuperCLEVR~\cite{DynSuperCLEVR}, the test introduces two additional settings: future prediction and counterfactual reasoning. Future prediction requires the model to infer the future motion state of an object based on its current speed and acceleration, such as ``\textit{How fast will the blue sphere be moving after 2 seconds?}''. Counterfactual reasoning, on the other hand, asks the model to reason about hypothetical scenarios where speed or acceleration changes, such as ``\textit{If the green cylinder's acceleration were doubled, when would it collide with the wall?}
The task of estimating velocity and acceleration requires the model to integrate displacement changes and temporal information within a 3D space while relying on an understanding of physical laws. The challenges of this task arise from two key issues: first, velocity and acceleration are multi-stage dynamic attributes. Estimating speed involves calculating displacement, followed by the estimation of acceleration, which creates a longer inference chain. Second, counterfactual reasoning and future prediction tasks require the model to reconstruct scenarios, involving complex ``what-if'' logic chains that are more difficult than simple factual queries.

\noindent\textbf{Spatial Imagination.}
In this paper, we define spatial imagination tasks as those require reasoning from a hypothetical viewpoint different from the input perspective, given a visual input. It is important to note that some works~\cite{OmniSpatial, Open3DVQA, ViewSpatial-Bench} refer to the camera viewpoint as the ``\textit{egocentric perspective}'', while others~\cite{SPHERE, COMFORT} refer to it as the ``\textit{allocentric} perspective.'' To avoid ambiguity, we use the term ``hypothetic perspective'' to distinguish it from the camera viewpoint, which is the perspective directly provided to the model. The inputs can be a single image, requiring the model to reason from a hypothetical perspective. For example, ``\textit{If the waiter is on the side of the vase, then who is on his left?}''~\cite{OmniSpatial} The SQA3D requires the model to first locate a hypothetical viewpoint in the video based on the description and then perform a question-answering task, such as ``Sitting at the edge of the bed and facing the sofa, can I go straight to the coffee table in front of me?'' The MINDCUBE~\cite{MINDCUBE} considers reasoning from limited views, aiming to assess the model's ability to mentally model space, such as ``If you move from view 1 to view 2, what is the furthest from you?'' There is a significant body of research~\cite{SITE, VSI-Bench, Spatial-MM, SPAR-Bench} in this area, some of which also accounts for spatial understanding differences across various viewpoints and language conventions~\cite{COMFORT}, or conducts reasoning from synthetic collections of images~\cite{SRBench}. The naming of spatial imagination tasks varies across studies, SQA3D~\cite{SQA3D} labeling it as situational question-answering, and SRBench~\cite{SRBench} referring to it as mental rotation. Spatial imagination tasks are not simply direct perception of static object attributes but require the model to construct spatial representations of objects in the mind and perform viewpoint transformations, involving dynamic operations on spatial structures. Thus, the challenge of spatial imagination lies in the need of the model to overcome the limitations of the input perspective, build accurate topological and spatial structural relationships from limited viewpoint inputs and imagine visual cues from the hypothetical perspective.

\begin{figure*}[!h]
\centering
\includegraphics[width=7in]{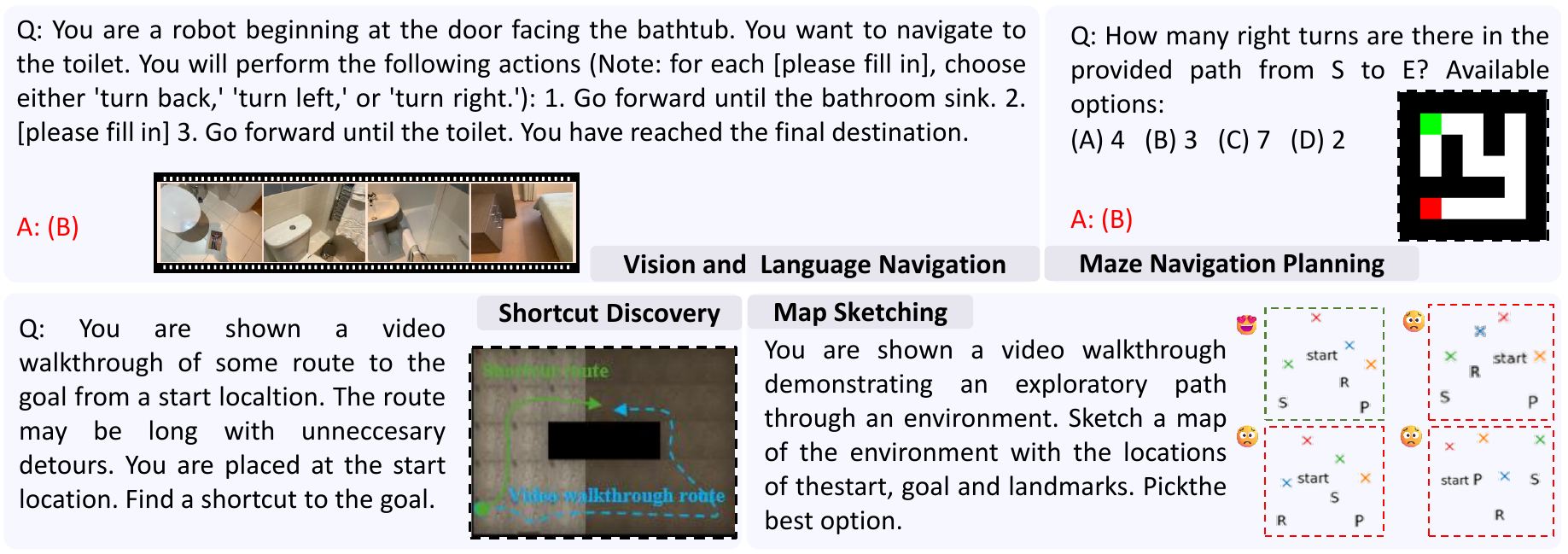}
\caption{Categorization of Spatial Planning Tasks.}
\label{fig:planning}
\end{figure*}

\subsection{Spatial Planning}
\label{sec:planning}
% SP定义：Receive previous observations from the environment, make planning prediction of the environment and next step
{Spatial planning aims to leverage previous observations of the environment to produce feasible actions and future predictions.}
{Compared with spatial perception and understanding, spatial planning goes a step further by transforming static recognition and relational reasoning into dynamic decision-making processes.}
% 根据prediction的种类，我们将SP分为environment planning; maze planning; Embodied planning
Given the different types of predictions, we divide spatial planning into three categories, \ie, environment planning, maze planning, and embodied planning.
Four representative tasks are shown in \reffig{fig:planning}.

\subsubsection{Environment Planning}

{Environment planning requires VLMs to generate a spatial representation of the surrounding environment and subsequently reason over it.
% use the representation to reason.
This process consists of two stages.
% It includes two stages. 
In the first stage, VLMs receive a video walking through the environment and build a representation based on it. In the second stage, they perform reasoning based on the constructed representation.
VLMs start to reason based on representation. Depending on the planning goals, this task can be divided into three settings.}

\noindent\textbf{Shortcut Discovery.}
After observing a video walkthrough from the start to the goal (with unnecessary steps), this task requires VLMs to plan a never-seen shortcut from the start to the goal~\cite{SPACE}.
% challenge点
The challenges of this task lie in (1) overall understanding of the environment, (2) interactive planning of actions given previous observations.

\noindent\textbf{Route Retracing.}
In SPACE~\cite{SPACE}, this task requires VLMs to retrace the route from the start after observing a video from the start to the goal. The route in the video is always the shortest path to the goal.
Similar to shortcut discovery, the challenge of this task also lies in the interactive planning of actions.

\noindent\textbf{Map Sketching.}
After observing a video walkthrough from the start to the goal, map sketching~\cite{SPACE} requires VLMs to pick the best map {representation} of the environment from multiple options.
% \footnote{multiple candidates? this is a multi-choice question}
The map contains landmarks, including the start and goal, with different spatial relationships.
The challenges of this task lie in (1) recognition of {critical} landmarks, (2) comprehension of spatial relationships among landmarks.
% \footnote{Can spatial relationships be planned????}

\subsubsection{Maze Navigation Planning}
Maze navigation planning~\cite{SPACE} involves navigating in a 2D grid world, from the start point to the exit point. Compared with general environment planning, maze planning features a clearly defined start and exit point, as well as a simpler environment. The key challenge lies in multi-hop reasoning while avoiding impassable locations.

Four types of elements are involved in this task, \ie, start point, exit point, impassable {points}, and passable points.
To describe these elements for VLMs, {two description modalities are adopted},
\ie, visual and textual. 
In the visual description, {elements are encoded in an image, with different colors indicating different element types}.
In the textual description, {elements are encoded in ASCII format, with different characters representing different element types}.

Maze navigation problems are often formulated as multiple-choice questions. Depending on the input modality, maze navigation planning can be categorized into three settings: {(1) text only, where mazes are represented in ASCII code~\cite{spatialeval}, (2) visual-only, where mazes are depicted as colored images~\cite{vot,SPACE,spatialeval}, and (3) text-visual, where both ASCII codes and colored images are provided~\cite{spatialeval}.}

\subsubsection{Embodied planning}
% 具身规划期望agent使用多模态数据理解环境 然后规划+自动take actions
Embodied planning expects agents to leverage multi-modal data to {perceive the surrounding environment, generate plans, and take actions in an autonomous manner.}
% 常规的 embodied planning是goal-orientied测评的 即 根据任务 和 任务完成情况 进行测评，例如在VLN任务中的success rate and average distance to goal
% General embodied planning tasks are designed for a task. 
{General embodied planning tasks are designed around completing a concrete goal.}
For example, in vision language navigation~\cite{wang2024vision}, agents are required to {reach a specific location following given instructions, with metrics like success rate and average distance to goal used to measure planning accuracy.}
% 在 VSR的任务下，大部分是要求VLMs来回答MCQ，其中MCQ的选项是由action space构成的
{By contrast, instead of physically navigating to a goal, embodied planning tasks in VSR require VLMs to solve multi-choice questions, with each option corresponding to possible actions.}
Depending on the action space, embodied planning in VSR includes vision-and-language navigation and mobile manipulation, both formulated as multi-choice questions.

\noindent\textbf{Vision-and-Language Navigation}
Vision-and-Language Navigation (VLN) in VSR requires agents to answer multi-choice questions while navigating through the environment.  
In VSI-Bench~\cite{VSI-Bench}, the questions are instructions with a cloze (fill-in-the-blank) format, and the answer options correspond to possible actions during the navigation process.  
In OmniSpatial~\cite{OmniSpatial}, each question includes the start point, end point, and orientation, while the options represent candidate navigation actions.  
OmniSpatial~\cite{OmniSpatial} also introduces a dynamic analysis setting, where question-answer pairs represent intermediate properties encountered during navigation, such as the number of doors passed.

\noindent\textbf{Mobile Manipulation.}
In VSR, mobile manipulation tasks ask agents to select the correct option that combines navigation with environment interaction, guided by the provided goals.
A representative benchmark for this task is SpaCe-10~\cite{gong2025space}, which focuses on room tidying tasks and formulates each choice to reflect aspects of the task flow, including navigation paths, goals, goal attributes, and the corresponding actions to be executed.

\section{Experiment}
\label{our eval}

\begin{figure*}[!h]
\centering
\includegraphics[width=7in]{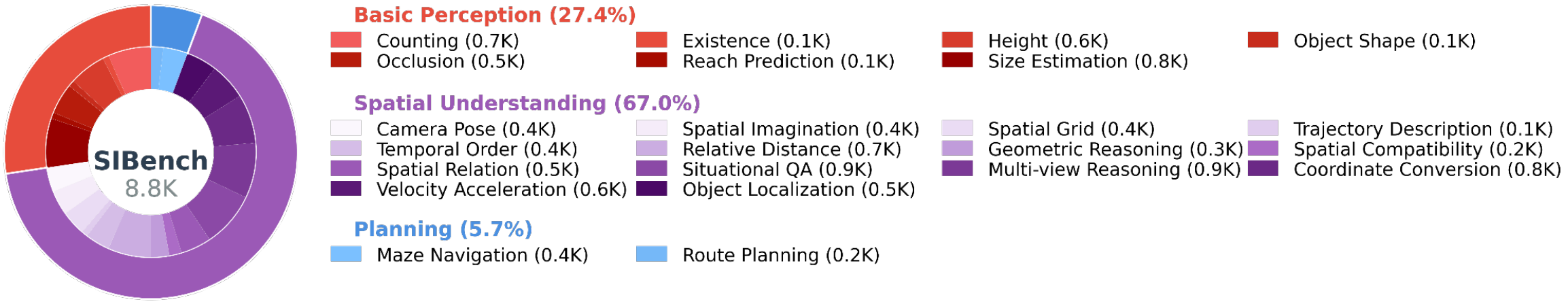}
\caption{\textbf{Statistical data of SIBench}. SIBench comprises VSR tasks across three cognitive levels, with a total of 8.8K data samples and 23 task settings.}
\label{fig:settings}
\end{figure*}
In this chapter, we first introduce the process of constructing the test data in the \refsection{sec:benchmark_construction}, including data collection, sampling criteria, and organization methods. Then, we use this test data to evaluate existing models and provide analysis and discussion based on the results in the \refsection{sec:results}.

\subsection{Benchmark Construction}
\label{sec:benchmark_construction}
The current benchmarks for VSR exhibit several issues, such as a lack of systematic and comprehensive task design, the presence of low-quality samples, a lack of human-annotated ground truth for some samples, and insufficient task rationale. 
To address these challenges, we survey existing open-source VSR benchmarks, aiming to integrate high-quality and diverse data. We ultimately develop SIBench, an evaluation benchmark that includes 23 task settings spanning 3 cognitive levels. 
% \reftab{tab:benchmark_comparison} presents a comparison of current benchmarks with SIBench.

\noindent{\textbf{Quality.}}
In our data quality assurance process, we prioritize datasets with manual annotations. 
For instance, when facing with a choice between manually annotated data and model-generated data for the same task, we consistently opt for the former. Only in the absence of manually annotated test data do we consider using semi-automatically annotated data that has undergone human review. Additionally, we filter out ``\textit{images}'' composed of emoticons and text to ensure that spatial information is derived from genuine visual input.

\noindent{\textbf{Diversity.}}
To construct a comprehensive evaluation dataset, we investigate nearly 20 open-source benchmarks, which we consolidate into 23 high-level task categories (see \reffig{fig:radar} for benchmark statistics). These categories span three cognitive levels: basic perception, spatial understanding, and planning. 
%It is important to note that under our classification method, a single task category may encompass several sub-tasks. For instance, the situated question-answering task includes distance measurement, spatial relationship determination, and camera posture estimation. 
For each task category, we strive to enhance the diversity of the test data. For example, for object size estimation, the data is sourced from three different benchmarks: SPHERE-VLM~\cite{SPHERE}, VSI-Bench~\cite{VSI-Bench}, and STI-Bench~\cite{STI-Bench}. Our test data includes three types of input formats: single images, multi-view images, and video data. The questions are presented in three formats: multiple-choice, true/false, and video data. 
%Furthermore, the same capability is tested in multiple formats. 
Details regarding the SIBench are available in \reffig{fig:radar} and \reffig{fig:settings}.
%We believe that this diversity in data formats and testing tasks enables a thorough assessment of a model's VSR capabilities.

\subsection{Evaluation}
\label{sec:results}
\begin{table*}[!t]
\centering
\caption{Performance Evaluation of Different Models on SIBench. Bold is the best, and underlining is the second best.}
\label{tab:model_performance}
% 使用 tabularray 环境来创建表格
\begin{tblr}{
  colspec = {lccccc}, % 定义列的格式
  row{1} = {font=\bfseries}, % 将第一行设置为粗体
  % row{2} = {bg=gray!25},     % 将第二行背景设置为50%的灰色
  % row{2, 3,4,5,6} = {bg=gray!25},   % 将第三和第四行背景设置为25%的灰色
  % row{6} = {bg=gray!15},   % 将第三和第四行背景设置为25%的灰色
  % hlines, % 添加所有默认的水平线
  hline{1,Z} = {wd=0.8pt}, % 将第一条和最后一条线加粗 (模拟 toprule/bottomrule)
  hline{2} = {wd=0.5pt}, % 将第二条和第五条线设为标准粗细 (模拟 midrule)
}
% 表格内容
Models & Rank & Overall  & Basic Perception & Spatial Understanding & Planning \\
% Random Choices &-& 0. &0.& 0. &0.25 \\
% GPT-4o-2024-11-20\textsuperscript{$\dagger$} & 4 & 0.4857  & 0.6339 & 0.4298 & 0.3375 \\
% Doubao-Seed-1.6-Vision & 3 & 0.6216 & 0.6963 & 0.5922& 0.65\\
% % GPT-5 -- low \textsuperscript{$\dagger$}  & 2 & \underline{0.6256} & 0.6835 & 0.5903 & 0.6875 \\
% Gemini-2.5-Pro\textsuperscript{$\dagger$}& 2 & {0.6295}  & 0.7317 & 0.5827 & 0.675 \\
% GPT-5 & 1 & 0.6906 & 0.7248 & 0.6487 & 0.775 \\
Random Choices &-& 0.3091 &0.4182& 0.2807 &0.25 \\
Qwen2.5-VL-7B~\cite{bai2025qwen2}  & 9 & 0.4712 & 0.5280 & 0.4624 & 0.3442 \\
Qwen2.5-VL-72B~\cite{bai2025qwen2}   &  4 & 0.5447  & 0.5793 & 0.5430 & 0.4161 \\
LLaVA-OneVision-7B~\cite{li2024llava}& 8 & 0.4850 & 0.5640 & 0.4708 & 0.3355 \\
LLaVA-OneVision-72B~\cite{li2024llava} & 7 & 0.5207 & {0.6086} & 0.5022 & 0.3922 \\
GPT-4o-mini~\cite{hurst2024gpt}       &  11 & 0.4278 & 0.5505 & 0.3981 & 0.3050 \\
Gemini-2.5-Flash~\cite{team2023gemini}  &  10  & 0.4389 & 0.5422 & 0.3942 & {0.6100} \\
InternVL-2.5-78B-MPO~\cite{bai2025qwen2} & 5 & 0.5338  & 0.5886 & 0.5165 & 0.5338 \\
InternVL-3-78B~\cite{zhu2025internvl3}     & 3 & 0.5481 & 0.6141 & 0.5344 & 0.4488\\
InternVL-3.5-38B~\cite{wang2025internvl3}  & 6 & {0.5252} & 0.5726 & {0.5134} & 0.4815 \\
Gemini-2.5-Pro~\cite{team2023gemini}     & 2  & \underline{0.5883} & \underline{0.6425} & \underline{0.5559} & \textbf{0.8017} \\
GPT-5~\cite{gpt5}     & 1  & \textbf{0.6341} & \textbf{0.6934} & \textbf{0.6158} & \underline{0.6296} \\
\end{tblr}
\end{table*}
\begin{table*}[!t]
\centering
\caption{Performance Evaluation of Different Models on the SIBench-mini. Bold is the best, and underlining is the second best.}
\label{tab:model_performance-mini}
% 使用 tabularray 环境来创建表格
\begin{tblr}{
  colspec = {lccccc}, % 定义列的格式
  row{1} = {font=\bfseries}, % 将第一行设置为粗体
  % row{2} = {bg=gray!25},     % 将第二行背景设置为50%的灰色
  % row{2, 3,4,5,6} = {bg=gray!25},   % 将第三和第四行背景设置为25%的灰色
  % row{6} = {bg=gray!15},   % 将第三和第四行背景设置为25%的灰色
  % hlines, % 添加所有默认的水平线
  hline{1,Z} = {wd=0.8pt}, % 将第一条和最后一条线加粗 (模拟 toprule/bottomrule)
  hline{2} = {wd=0.5pt}, % 将第二条和第五条线设为标准粗细 (模拟 midrule)
}
% 表格内容
Models & Rank & Overall  & Basic Perception & Spatial Understanding & Planning \\
% Random Choices &-& 0. &0.& 0. &0.25 \\

LLaVA-OneVision-72B~\cite{li2024llava} & 6 & 0.5252 & 0.6136  & 0.5065 & 0.3968 \\

Qwen2.5-VL-72B~\cite{bai2025qwen2} & 7 & 0.5168 & 0.5930 & 0.4958 & 0.4743 \\

InternVL-3.5-38B~\cite{InternVL3} & 5 & 0.5355 & 0.6113 & 0.5089 & 0.4878 \\
GLM4.5-V-106B-A12B~\cite{hong2025glm} & 4 & 0.5822 & 0.6936 & 0.5404 & 0.5125 \\
% Qwen2.5-VL-72B &&&&& \\
% GPT-4o-2024-11-20\textsuperscript{$\dagger$} & 6 & 0.4857  & 0.6339 & 0.4298 & 0.3375 \\
Doubao-Seed-1.6-Vision~\cite{guo2025seed1} & 3 & 0.6216 & 0.6963 & \underline{0.5922}& 0.65\\
% GPT-5 -- low \textsuperscript{$\dagger$}  & 2 & \underline{0.6256} & 0.6835 & 0.5903 & 0.6875 \\
Gemini-2.5-Pro~\cite{team2023gemini} & 2 & \underline{0.6295}  & \textbf{0.7317} & 0.5827 & \underline{0.675} \\
GPT-5~\cite{gpt5} & 1 & \textbf{0.6906} & \underline{0.7248} & \textbf{0.6487} & \textbf{0.775} \\
\end{tblr}
\end{table*}

\subsubsection{Setup.}
We perform a detailed comparative analysis of state-of-the-art proprietary and open-source models. Our evaluation of proprietary models includes GPT-5~\cite{gpt5}, Gemini-2.5-Pro~\cite{team2023gemini}, Gemini-2.5-Flash~\cite{team2023gemini} and GPT-4o-mini~\cite{hurst2024gpt}. The primary open-source models under consideration are Qwen2.5-VL-72B~\cite{bai2025qwen2}, LLaVA-OneVision-72B~\cite{li2024llava} and InternVL-3-78B~\cite{zhu2025internvl3}. Consistency is maintained throughout our experimental design. All models process images (both single and multi-view) at an identical resolution. For video inputs, we apply a standard procedure of uniformly sampling 30 frames. Furthermore, we utilize specific prompt templates for each QA category; for example, multiple-choice questions are prefaced with "Select from the following choices," and numerical questions are prompted with "Answer using a number." 

By default, we do not employ Chain-of-Thought (CoT) prompting. To handle any non-formatted outputs, we utilize Gemini-1.5-Flash~\cite{team2023gemini} for post-processing to standardize the answers. The evaluation metric depends on the question type. For True/False (TF) and Multiple-Choice Questions (MCQ), we directly compare the post-processed predictions against the ground truth. For questions requiring a numerical answer, we adopt the Mean Relative Accuracy ($\mathcal{MRA}$) metric from VSI-Bench~\cite{VSI-Bench}, which is calculated as follows:
    \begin{equation} \label{eq:mra}
\mathcal{MRA} = \frac{1}{10} \sum_{\theta \in C} \mathbb{1} \left( \frac{|\hat{y} - y|}{y} < 1 - \theta \right),
\end{equation}
where $\hat{y}$ is the model's prediction, $y$ is the ground truth, $\theta$ is a confidence threshold from the set $C = \{0.5, 0.55, \dots, 0.95\}$.

To streamline the evaluation process, we construct a mini version named SIBench-mini comprising 40 randomly selected samples from each task setting for evaluating Gemini-2.5-Pro~\cite{team2023gemini}, Doubao-Seed-1.6-Vision~\cite{guo2025seed1}, GLM4.5-V-106B-A12B~\cite{zeng2025glm}, and GPT-5~\cite{gpt5}. This subset is as comprehensive as the full benchmark in terms of task settings, but at a much lower testing cost.

\begin{table*}[t]
\centering
\caption{Performance evaluation of different models on visual spatial reasoning tasks. Bold is the best, and underlining is the second best.}
\label{tab:performance_settings}
\resizebox{\textwidth}{!}{
\begin{tblr}{
colspec = {Q[l, m] Q[l, m] *{4}{Q[c, m]}},
hline{1,Z} = {wd=0.8pt},% Top and bottom rules
hline{3} = {1-6}{wd=0.4pt},% cmidrule under Models
hline{2, 10, 24, 26} = {wd=0.4pt},% Main section separators
}
% Header
\SetCell[r=2]{c} Settings & \SetCell[r=2]{c} & \SetCell[c=4]{c} Models \\
& & Qwen2.5-VL-72B~\cite{bai2025qwen2} & LLaVA-OneVision-72B~\cite{li2024llava} & GPT-5~\cite{gpt5} & Gemini-2.5-Pro~\cite{team2023gemini} \\
% Basic Perception
\SetCell[r=7]{c, font=\bfseries} Basic Perception & Reach Prediction & \textbf{0.6750} & \underline{0.6000} & 0.5750 & 0.5500 \\
& Height & 0.6000 & 0.6466 & \underline{0.6800} & \textbf{0.7133} \\
& Existence & \textbf{1.0000} & \underline{0.9500} & 0.9250 & 0.9000 \\
& Occlusion & 0.5580 & \underline{0.6660} & 0.6460 & \textbf{0.7300} \\
& Object Shape & 0.3538 & 0.3077 & \textbf{0.5154} & \underline{0.4154} \\
& Counting & \underline{0.5504} & 0.5084 & \textbf{0.7408} & 0.4664 \\
& Object Size Estimation & \underline{0.6085} & 0.6045 & \textbf{0.7645} & 0.5968 \\
% Spatial Understanding
\SetCell[r=14]{c, font=\bfseries} Spatial Understanding & Spatial Compatibility & \underline{0.6028} & 0.5841 & {0.5607} & \textbf{0.7103} \\
& Coordinate Conversion & 0.6513 & 0.5950 & \textbf{0.7788} & \underline{0.6900} \\
& Trajectory Description & \textbf{0.6026} & \underline{0.5513} & 0.3205 & 0.3333 \\
& Geometric Reasoning & 0.2421 & 0.2579 & \textbf{0.5278} & \underline{0.3730} \\
& Spatial Imagination & \underline{0.2900} & 0.2650 & \textbf{0.4575} & 0.2700 \\
& Spatial Grid & {0.7875} & 0.5175 & \underline{0.7900} & \textbf{0.9975} \\
& Temporal Appearance Order & \textbf{0.5455} & \underline{0.3636} & 0.2957 & \textbf{0.5455} \\
& Multi-View Reasoning & 0.4072 & 0.4316 & \textbf{0.5854} & \underline{0.4687} \\
& Situational QA & 0.5295 & 0.5128 & \textbf{0.6055} & \underline{0.5474} \\
& Velocity Acceleration & \underline{0.4605} & 0.4403 & \textbf{0.5746} & 0.4034 \\
& Relative Distance & \underline{0.5675} & 0.5037 & \textbf{0.6828} & {0.5459} \\
& Camera Pose & \underline{0.4738} & 0.3169 & \textbf{0.5000} & 0.4041 \\
& Spatial Relation & \textbf{0.7978} & \underline{0.7870} & 0.6929 & 0.6660 \\
& Object Localization & 0.7134 & \underline{0.7570} & \textbf{0.7727} & 0.7041 \\
% Planning
\SetCell[r=2]{c, font=\bfseries} Planning & Maze Navigation & 0.4350 & 0.3975 & \underline{0.7725} & \textbf{0.8625} \\
& Route Planning & 0.2881 & \underline{0.3559} & {0.3351} & \textbf{0.3898} \\
% Overall
\SetCell[c=2]{c, font=\bfseries} Overall & & 0.5447 & 0.5207 & \textbf{0.6341} & \underline{0.5883} \\
\end{tblr}%
}
\end{table*}

\subsubsection{Main Results.}\label{sec:main_results}
% modify the result
\textbf{Comparison of Models}.
The experimental results indicate that leading proprietary models establish the highest tier of performance (see \reftab{tab:model_performance}). Specifically, GPT-5~\cite{gpt5} achieves the best performance across the three levels of visual-spatial reasoning included in SIBench and SIBench-mini.
%GPT-5 is particularly noteworthy, achieving the best scores on both the full test set (Overall: 0.5883) and the more uniformly distributed mini test set (Overall: 0.5928), followed by GPT-5 (Overall: 0.5796). 
Secondly, while a performance gap persists between open-source models and their leading proprietary counterparts, the former exhibit strong competitiveness against lightweight proprietary models. Although representative open-source models, such as InternVL-3.5-38B~\cite{wang2025internvl3} (overall score: 0.5252) and Qwen2.5-VL-72B~\cite{bai2025qwen2} (overall score: 0.5114), lag behind the leading models, their overall performance still outperforms lightweight models like GPT-4o-mini (0.4278) and Gemini-2.5-Flash~\cite{team2023gemini} (0.4389). Furthermore, we observe that GPT-5 and Gemini-2.5-Pro holds a significant lead in the dimensions of spatial understanding and planning, which may imply that these premier models possess superior abstract reasoning capabilities.
On SIBench-mini, similar patterns are observed. Specifically, as shown in Table~\ref{tab:model_performance-mini}, the proprietary model GPT-5~\cite{gpt5} achieves the best performance, followed by Gemini-2.5-Pro and Doubao-Seed-1.6-Vision~\cite{guo2025seed1}. In contrast, other open-source models, such as GLM4.5-V-106B-A12B~\cite{zeng2025glm} and InternVL-3.5-38B~\cite{InternVL3}, still exhibit a noticeable performance gap.

\begin{figure*}[!h]
\centering
\includegraphics[width=7in]{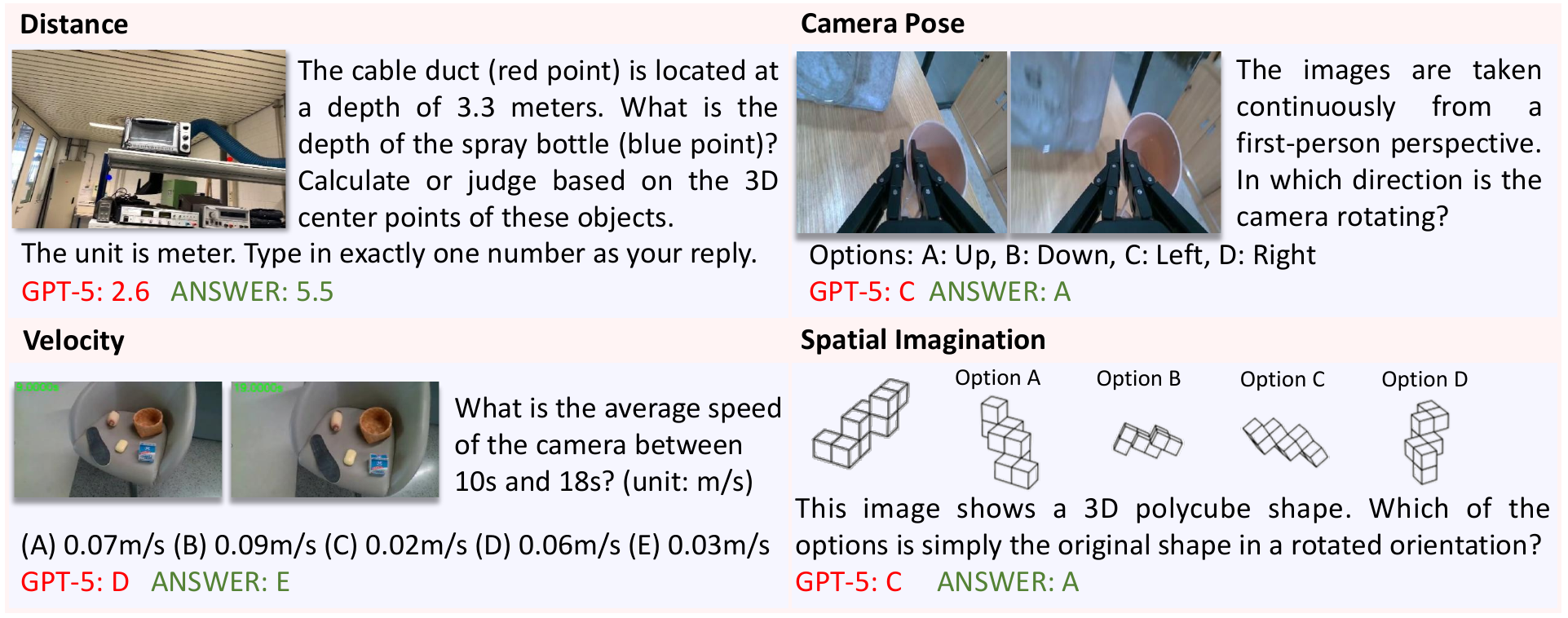}
\caption{\textbf{Failure Cases in Spatial Reasoning Tasks}, including quantitative reasoning (upper-left), multi-view reasoning (upper-right), non-static logical inference (bottom-left), and spatial imagination (bottom-right).}
\label{fig:failure_case}
\end{figure*}

\noindent\textbf{Comparison of Various Settings.}
Our evaluation reveals a stark contrast in the visual spatial capabilities of current models (see \reftab{tab:performance_settings}, \reftab{tab:performance_supp} and \reftab{tab:mini_performance_supp}). They exhibit high proficiency in fundamental perceptual tasks, such as identifying object existence, understanding occlusion, and judging qualitative spatial relationships, indicating strong static feature extraction. However, this proficiency does not extend to higher-level cognitive reasoning. 
% As shown in figure~\ref{radar}, four core deficiencies are identified (see Appendix for specific experimental data). (1) The models lack quantitative reasoning, failing to accurately estimate physical quantities like distance and size. (2) These models struggle with multi-view input problems, such as tasks involving reasoning about camera pose transformations.(3) Moreover, they also show a performance decline when dealing with temporal tasks.This suggests their logical inference is fragile when dealing with non-static inputs. (4) The most profound weakness is a near-total absence of spatial imagination. 
\reffig{fig:radar} and \reffig{fig:failure_case} highlight four core deficiencies in the models, which are substantiated by detailed experimental data in the Appendix (see \reftab{tab:performance_supp} and \reftab{tab:mini_performance_supp}).
First, the models exhibit poor quantitative reasoning, often failing to accurately estimate physical quantities such as distance and size (see the upper-left part of \reffig{fig:failure_case}). This limitation likely arises from their reliance on coarse visual heuristics rather than precise metric representations.
Second, they struggle with multi-view reasoning, particularly in tasks requiring the inference of camera pose transformations (see the upper-right part of \reffig{fig:failure_case}). This weakness may stem from the lack of explicit geometric constraints in training.
Third, their performance degrades significantly on temporal tasks. This fragility in handling non-static inputs suggests their underlying logical inference is unreliable (see the bottom-left part of \reffig{fig:failure_case}). A plausible reason is that these models primarily process frames independently, failing to capture coherent temporal dynamics.
Finally, and perhaps most critically, the models exhibit a near-total absence of spatial imagination. All models failed on tasks requiring the creation and manipulation of a mental model (see the bottom-right part of \reffig{fig:failure_case}). 
% First, the models exhibit poor quantitative reasoning, often failing to accurately estimate physical quantities such as distance and size (see the upper-left part of \reffig{fig:failure_case}).
% Second, they struggle with multi-view reasoning, particularly in tasks requiring the inference of camera pose transformations (see the upper-right part of \reffig{fig:failure_case}).
% Third, their performance degrades significantly on temporal tasks. This fragility in handling non-static inputs suggests their underlying logical inference is unreliable (see the bottom-left part of \reffig{fig:failure_case}).
% Finally, and perhaps most critically, the models  a near-total absence of spatial imagination. All models failed on tasks requiring the creation and manipulation of a mental model (see the bottom-right part of \reffig{fig:failure_case}). 
This exposes a fundamental limitation: they excel at interpreting visible information but cannot conceive of what is absent, functioning as observers rather than reasoners. In short, while models have become adept at "seeing," their ability to perform cognitive "thinking", particularly spatial imagination and precise inference, remains nascent.
\section{Discussion}
\label{sec:discussion}
In this section, we first summarize the challenges that VLMs face in Visual Spatial Reasoning tasks in \refsection{sec:challenges}, and then propose several potential solutions in \refsection{sec:solutions}.

\begin{figure*}[!h]
\centering
\includegraphics[width=7in]{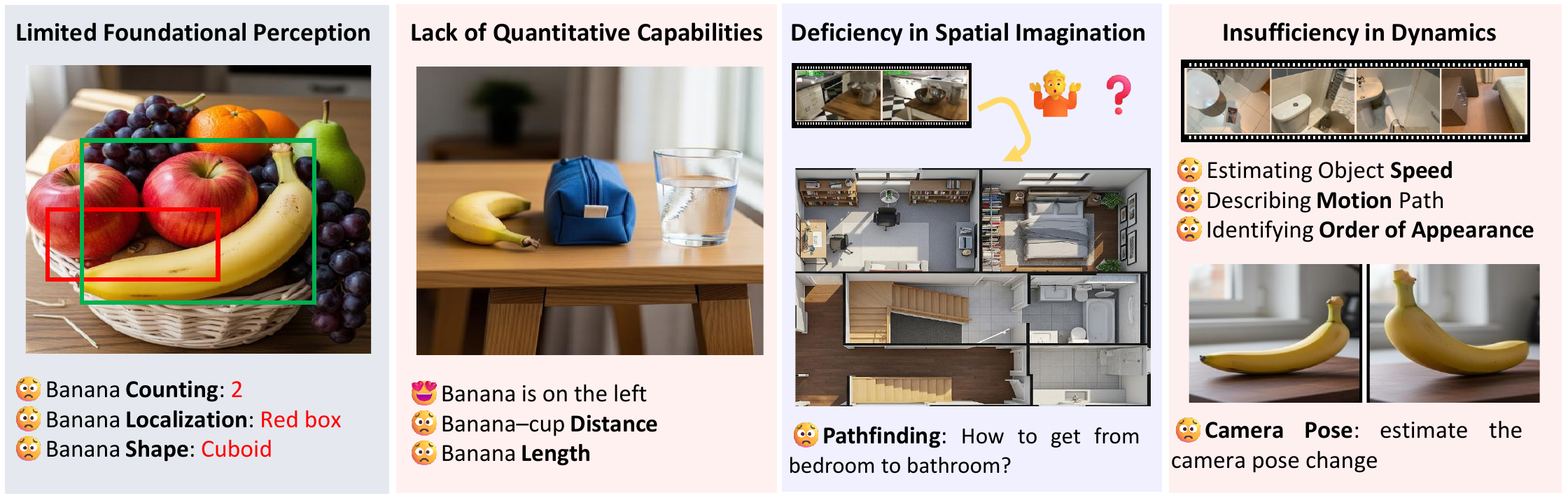}
\caption{\textbf{Challenges for VLMs in VSR tasks}. From left to right, examples illustrate VLM deficiencies in foundational perception, quantitative reasoning, spatial imagination, and dynamic reasoning. Detailed examples can be found in the supplementary materials.}
\label{fig:challenges}
\end{figure*}

\subsection{Challenges}
\label{sec:challenges}
While VLMs have advanced significantly in multimodal understanding, they face critical challenges in complex VSR. These challenges constrain their performance on benchmarks and impede real-world deployment in fields like robotics and autonomous driving. Based on the evaluation of existing models, we summarize these challenges into the following four main aspects (see \reffig{fig:challenges}).

\noindent{\textbf{Limited Robustness of Foundational Perception.}} Despite the strong overall perceptual capabilities of VLMs, their foundational perception capabilities often lack robustness, particularly in complex or atypical scenarios. These uncertainties at the perceptual level can directly undermine the reliability of any subsequent spatial reasoning. For example, \textit{Counting} remains a significant challenge, especially in dense or occluded scenes, while precise \textit{Object Localization} and fine-grained \textit{Shape Recognition} are also far from perfect. When deviations occur at these fundamental perceptual stages, such as misidentifying the number or location of objects, any reasoning based on this flawed information becomes invalid. This "error accumulation" effect highlights that even if a model possesses the potential for advanced reasoning, its performance is ultimately constrained by the bottlenecks within its foundational perception capabilities.

\noindent{\textbf{Lack of Precise and Quantitative Capabilities.}} VLMs exhibit inherent shortcomings in precision and quantification, a limitation that may stem from their reliance on learning from human language, which is often ambiguous when describing the physical world. Consequently, while these models excel at processing qualitative spatial relationships, such as "\textit{A is to the left of B,}" they exhibit significant deficiencies in tasks requiring quantitative geometric computation. This deficiency is empirically validated by benchmark results. As illustrated in \reffig{fig:radar}, existing models consistently underperform on tasks that demand precise numerical regression, including \textit{Coordinate Conversion}, \textit{Distance Prediction}, and \textit{Object Size Estimation}. This difficulty in mapping visual features to precise physical coordinates or metric units severely restricts the application potential of VLMs in high-precision scenarios, such as robotic grasping and the accurate placement of content in Augmented Reality (AR).

\noindent{\textbf{Deficiency in Spatial Imagination and 3D Reconstruction.}} Primarily relying on 2D images as input, VLMs generally lack the ability to "imagine" or reconstruct a complete 3D world from limited perspectives. This deficiency hinders their ability to comprehend the physical properties of objects and the spatial relationships involving occluded parts. Evaluation results intuitively reflect this issue. For instance, in \textit{Situational Question and Answering} tasks, models struggle to infer the complete geometry of unseen objects. Similarly, in tasks that require a mental simulation of spatial connectivity, such as \textit{Route Planning} and \textit{Maze Navigation}, VLMs without a robust 3D world model also perform poorly. This absence of capability prevents the models from constructing a coherent and global scene representation, which is a major obstacle to high-level spatial cognition.

\noindent{\textbf{Insufficiency in Dynamic-Temporal and Cross-View Reasoning.}} The real world is inherently dynamic and variable, necessitating that an intelligent agent integrate information across time and viewpoints to form a coherent spatial cognition. However, current VLMs exhibit notable insufficiencies in this area. The \reffig{fig:radar} highlights the models' weaknesses in tasks such as \textit{Multi-View Reasoning}, \textit{Velocity \& Acceleration}, and \textit{Trajectory Description}. This suggests that VLMs struggle both to track the spatio-temporal state of objects effectively and to align and fuse information from disparate camera viewpoints. This deficiency in processing dynamic and multi-view information poses a critical barrier to the advancement of VLMs into high-level applications requiring continuous interaction with dynamic environments, such as robotics and autonomous driving.

\subsection{Potential Solutions}
\label{sec:solutions}

\noindent{\textbf{Constructing Higher-Quality and More Diverse Training Data.}} A foundational solution to the current challenges for VLMs is the construction of higher-quality and more diverse training data. The core of this strategy lies in increasing data diversity, not only in terms of sources (\eg{, real-world images, videos, synthetic data}) but more critically, in the breadth of task settings, as different tasks directly correspond to distinct foundational spatial abilities. For instance, relational reasoning in VSR can train a model's understanding of relative positions; maze navigation or route planning tasks can foster a global spatial-topological cognition; and simulated robotic manipulation tasks demand more fine-grained localization and judgement of physical affordance. An ideal training strategy should therefore combine diverse task settings with high-quality structured annotations. Leveraging large-scale synthetic data with perfect ground truth from simulation engines like Blender and CARLA is also an essential component of this approach.

\noindent{\textbf{Incorporating 3D-Aware and Fine-Grained Perception Tasks in Pre-training Stage.}} Incorporating 3D-aware and fine-grained perception tasks into the pre-training phase is crucial for endowing VLMs with robust spatial imagination and perceptual robustness. To facilitate a model's understanding of the 3D world, large-scale synthetic 3D datasets (\eg{, ShapeNet, Objaverse}) can be utilized for tasks such as novel view synthesis or 3D shape reconstruction. These tasks compel the model to learn an internal, view-invariant object representation. Concurrently, to enhance the robustness of foundational perception, a multi-task learning paradigm can be employed where the visual encoder learns auxiliary tasks like depth estimation, surface normal prediction, and instance segmentation alongside its primary vision-language alignment objective. This multi-task setup acts as a powerful regularizer, compelling the visual backbone to learn features that are generalizable across various geometries and semantics, thereby providing a more reliable input for higher-level reasoning and effectively mitigating the "error accumulation" effect.

\noindent{\textbf{Towards Advanced Unified Spatiotemporal Architectures.}} Most existing VLM architectures are designed to process static, discrete images, creating a fundamental disconnect with the continuous, dynamic, and multi-view perception of an intelligent agent in the real world. Therefore, a key direction for architecture evolution is the development of a unified spatiotemporal architecture oriented towards Embodied AI. Such an architecture would no longer treat space and time as separate dimensions but would instead perceive the world as a continuous four-dimensional spacetime stream. Its core objective is to enable the model to integrate information from different times and viewpoints, forming a dynamic, consistent, and ego-centric understanding of the scene. This path aims to transcend the model's limitation as a passive '\textit{bystander}' and empower it with the ability to perceive and predict the world as an active '\textit{agent}.' This evolution is an essential pathway toward advanced applications such as robotics and autonomous driving.

\subsection{Applications}
As this review has elaborated, VSR represents not only a theoretical challenge but also the critical capability for translating digital intelligence into real-world action. The profound impact of VSR is most vividly exhibit across different frontier domains, such as autonomous driving, embodied robotics, and augmented reality.

\noindent{\textbf{Embodied Intelligence.}} For embodied AI and robotics, VSR acts as the crucial bridge connecting abstract instructions to concrete physical actions~\cite{VSI-Bench,krantz_vlnce_2020}. It enables agents to ground natural language commands (\eg{``place the cup in the center of the table''}) into precise 3D spatial coordinates and action sequences~\cite{wang2024vision}, thereby facilitating complex navigation~\cite{VSI-Bench,OmniSpatial} and manipulation tasks~\cite{gong2025space}. Although current models still exhibit deficiencies in geometric and physical commonsense reasoning, the rise of Vision-Language-Action (VLA) models~\cite{qu2025spatialvla,zheng2024tracevla}, particularly those incorporating internal reasoning mechanisms like the ``visual chain-of-thought''~\cite{Cot-VLA}, is advancing robotic intelligence toward a "think before you act" paradigm. This approach effectively decomposes complex tasks and improves the success rate of planning and execution.

\noindent{\textbf{Autonomous Driving.}} In the domain of autonomous driving, VSR serves as the cognitive cornerstone for safe navigation. It elevates a vehicle's perceptual capabilities from simple object recognition to deep understanding of the 3D structure of dynamic traffic scenes~\cite{zhou2024embodied,cao2024maplm,tian2025nuscenes}, the relative relationships between objects~\cite{choudhary2024talk2bev}, and their motion trajectories~\cite{sachdeva2024rank2tell}. This capability is a prerequisite for all advanced planning and decision-making. A central challenge in this field is bridging the gap between the 2D image-based reasoning abilities developed during model training and the need for precise, quantitative inference in the 3D physical world~\cite{zhou_vlmad_survey}. To this end, the field is shifting from imitation learning towards explicit reasoning, developing end-to-end models like EMMA~\cite{hwang2024emma} and neuro-symbolic hybrid frameworks such as Logic-RAG~\cite{kabir2025logic}, to build more robust and interpretable driving intelligence.

\section{Conclusion}
In this paper, we presents a systematic and comprehensive investigation into the domain of Visual Spatial Reasoning (VSR) for Vision-Language Models (VLMs). We begin by conducting a review of existing methodologies, analyzing them through input modalities, model architectures, training strategies, and inference mechanisms. Concurrently, we introduce a cognition-based hierarchical taxonomy that categorizes VSR tasks into three distinct levels: Basic Perception, Spatial Understanding, and Spatial Planning. Building upon this framework, we curate SIBench, a comprehensive benchmark that integrates nearly 20 open-source datasets across 23 task settings to facilitate a holistic evaluation of VLM capabilities. Our experiments on state-of-the-art models using SIBench reveal a critical gap between perceptual and cognitive abilities. While models exhibit proficiency in basic perception, they exhibit significant deficiencies in higher-order reasoning, particularly in tasks requiring precise numerical estimation, multi-view reasoning, dynamic temporal understanding, and spatial imagination. These findings underscore the substantial challenges that remain on the path to achieving spatial intelligence in artificial agents.

\ifCLASSOPTIONcaptionsoff
  \newpage
\fi

{\small
\bibliographystyle{IEEEtran}
\bibliography{main}

}
% --- 在参考文献之后添加附录 ---
\clearpage
\begin{appendices}
\section{Appendix Outline}
This supplementary material includes {6} aspects:
\begin{enumerate}
    \item Data Source: the distribution of the proposed SIBench.
    \item Adaptation of test data.
    \item Detailed Comparison of Various Settings.
    \item Failure case analysis.
    \item Timeline of Representative VSR Benchmarks.
    \item Comparison of Benchmarks for Visual Spatial Reasoning.
\end{enumerate}

\subsection{Data Source}

\begin{figure}[!h]
\centering
\includegraphics[width=3.5in]{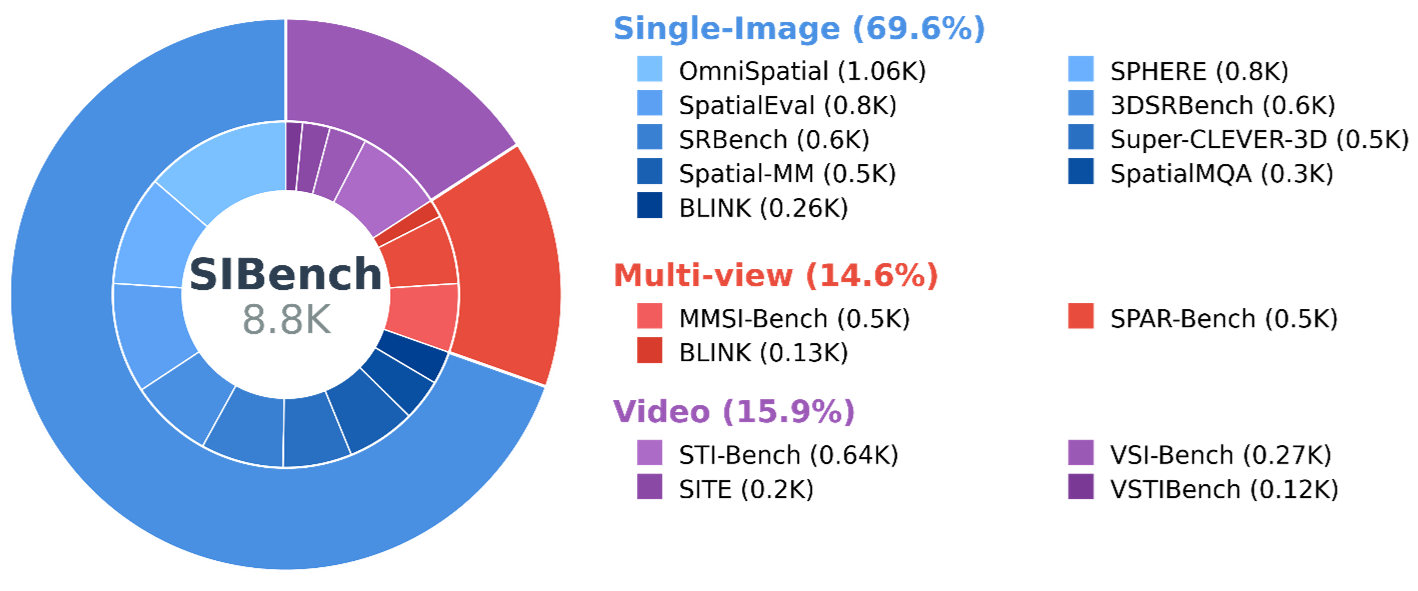}
\caption{\textbf{Data source of SIBench}. The distribution of SIBench across three input formats and nearly 20 open-source datasets.}
% Statistical data of task settings. The data is sourced from nearly 20 open-source datasets and covers three input formats
\label{fig:input_source}
\end{figure}

We construct SIBench by aggregating data from approximately 20 open-source datasets into three distinct input formats (see \reffig{fig:input_source}). The breadth of the data guarantees the comprehensiveness of the evaluation.

\subsection{Adaptation of test data}
\begin{figure}[!h]
\centering
\includegraphics[width=3.5in]{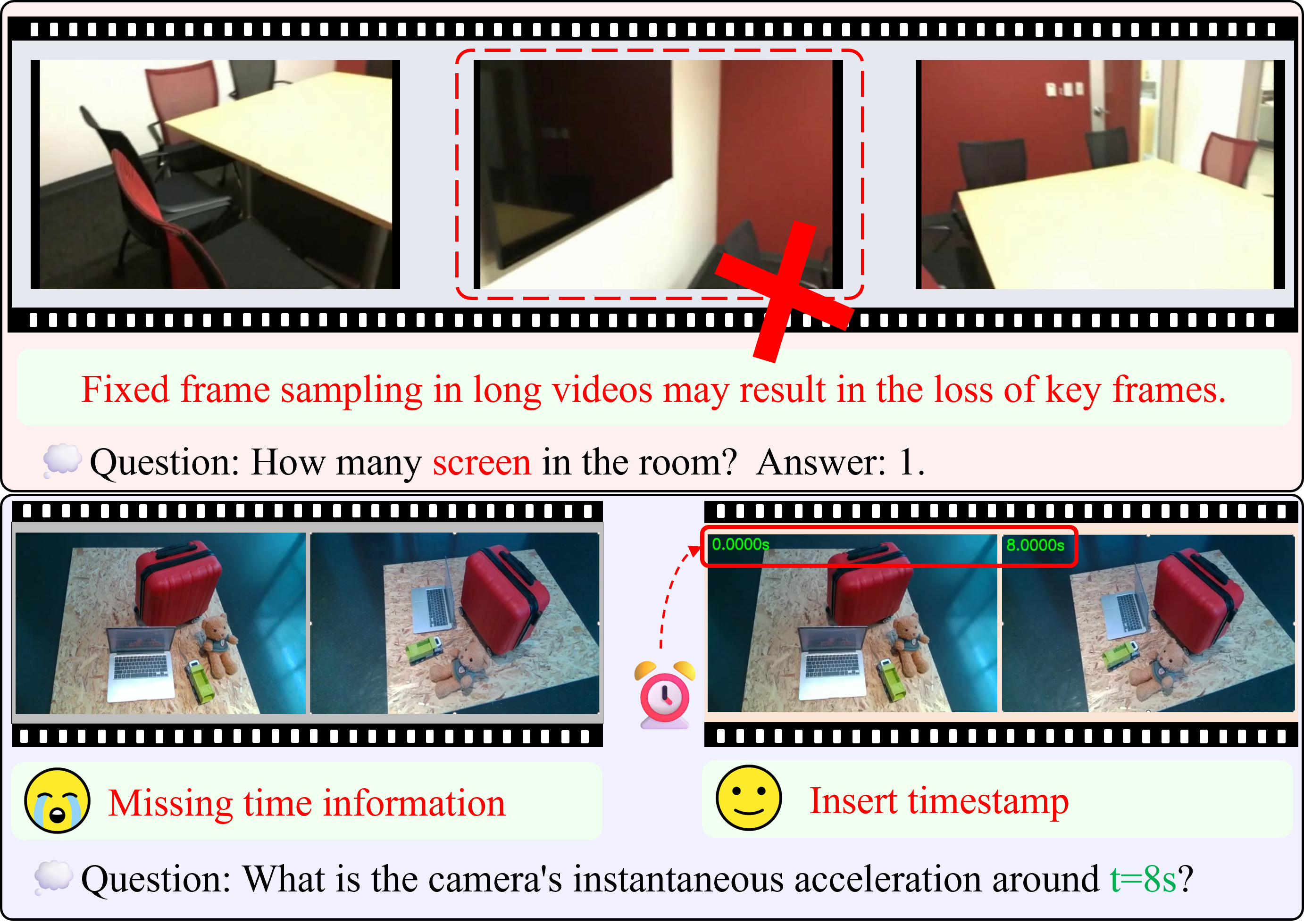}
\caption{\textbf{Issues with the test data}. i. (Upper) When the test video is too long, the larger sampling interval may cause some targets to be lost, leading to mismatches with the ground truth. ii. (Bottom) The variable sampling interval results in the loss of temporal information.}
\label{fig:quality}
\end{figure}

We conduct a thorough assessment of the task's reasonableness. In our video test data, for example, we observe that some videos are as long as five minutes, yet the testing protocol requires sampling only 30 frames for model input. This large sampling interval often leads to target loss, which adversely affects tasks such as object counting (see the upper part of \reffig{fig:quality}). To mitigate this, we restrict our video selections to a maximum duration of one minute. In another example, when a model is tasked with estimating velocity and acceleration from a video, the testing requirements mandate a fixed number of sampled frames. However, the video durations are variable, resulting in inconsistent sampling intervals and a loss of temporal information, which renders the test meaningless. Our solution is to add a timestamp to each frame, ensuring that effective temporal information is preserved regardless of variations in the sampling interval (see the bottom part of \reffig{fig:quality}).

\subsection{Detailed Comparison of Various Settings}
\label{sec:detail_comparision_of_various_setting}
% SIBench is an evaluation benchmark that includes 23 task settings. Detailed comparison of various settings on SIBench and SIBench-mini are shown in \reftab{tab:performance_supp} and \reftab{tab:mini_performance_supp}, respectively.
SIBench is a comprehensive benchmark consisting of 23 distinct task settings. The detailed results of various models on SIBench and SIBench-mini are shown in \reftab{tab:performance_supp} and \reftab{tab:mini_performance_supp}, respectively.
Models generally excel at basic perception tasks like identifying the existence of objects but struggle significantly with tasks requiring complex, abstract reasoning, such as spatial imagination, temporal ordering, and precise quantitative estimation. Overall, Gemini-2.5-Pro~\cite{team2023gemini} is the top one in the SIBench, while the GPT-5~\cite{gpt5} leads in the SIBench-mini.
\subsubsection{Areas of Model Strength}
In \reftab{tab:performance_supp} and \reftab{tab:mini_performance_supp}, models consistently perform well on tasks that rely on recognizing objects and simple relationships of objects.
On the existence task, {Qwen2.5-VL-72B~\cite{bai2025qwen2}} achieves a perfect score of {1.0}, and many others are above 0.9. This indicates that identifying whether an object is present in a scene is an easy task for current models.
On the Spatial Relation task, {InternVL-3-78B~\cite{zhu2025internvl3}} leads at {0.8402} and {Qwen2.5-VL-72B~\cite{bai2025qwen2}} also receives 0.8216. This strength suggests that models have a solid grasp of static relational ability.
\subsubsection{Areas of Model Weakness}
The models exhibit several key weaknesses, consistent with the analysis in \refsection{sec:main_results}.
% 感知
% 
First, models are weak at some tasks of Foundational Perception, such as {Object Size Estimation} shown in \reftab{tab:performance_supp} and \reftab{tab:mini_performance_supp}.
% 定量
% 
Second, models struggle with quantitative reasoning. This is evident in tasks {Relative Distance} in \reftab{tab:performance_supp} and \reftab{tab:mini_performance_supp}. The scores indicate that models like Gemini-2.5-Flash~\cite{team2023gemini} and InternVL-3-78B~\cite{zhu2025internvl3} rely on coarse visual clues rather than precise metric representations.
% 想象
% 
Third, models show a near-total absence of spatial imagination. On the Spatial Imagination task, models like GPT-4o-mini~\cite{hurst2024gpt} and InternVL-2.5-78B-MPO~\cite{zhu2025internvl3} receive almost the lowest scores. This indicates that models fail on tasks that require the creation and manipulation of a mental imagination. Models cannot visualize and reason about objects or scenarios that are not explicitly depicted, which is a key component of human-like spatial intelligence.
% 动态
Second, models are weak at inferring spatial properties from different viewpoints. This is shown by the low scores in {Multi-View Reasoning}, {Camera Pose}, and {Geometric Reasoning}. Models like Qwen2.5-VL-7B~\cite{bai2025qwen2}, LLaVA-OneVision-72B~\cite{li2024llava}, Gemini-2.5-Flash~\cite{team2023gemini} cannot intuitively understand how a camera's movement changes a scene's perspective, leading to poor performance. Similarly, models struggle with temporal appearance order, with LLaVA-OneVision-7B~\cite{li2024llava} scoring {0.0000}. This confirms that models' unreliable ability for non-static scenes.

\begin{table*}[t]
\centering
\caption{Performance evaluation of different models on visual spatial reasoning tasks.}
\label{tab:performance_supp}
\resizebox{\textwidth}{!}{
\begin{tblr}{
  colspec = {Q[l, m] Q[l, m] *{7}{Q[c, m]}},  % Changed from 10 to 7
hline{1,Z} = {wd=0.8pt},% Top and bottom rules
  hline{3} = {1-9}{wd=0.4pt}, % cmidrule under Models (changed from 1-12 to 1-9)
  hline{2, 10, 24, 26} = {wd=0.4pt},% Main section separators
}
% Header
\SetCell[r=2, c=2]{c} Settings & & \SetCell[c=7]{c} Models \\ % Changed from c=10 to c=7
& & Qwen2.5-VL-7B~\cite{bai2025qwen2} & LLaVA-OneVision-7B~\cite{li2024llava} & GPT-4o-mini~\cite{hurst2024gpt} & Gemini-2.5-Flash~\cite{team2023gemini} & InternVL-2.5-78B-MPO~\cite{zhu2025internvl3} & InternVL-3-78B~\cite{zhu2025internvl3} & InternVL-3.5-38B~\cite{zhu2025internvl3} \\
% Basic Perception
\SetCell[r=7]{c, font=\bfseries} Basic Perception & Reach Prediction & 0.6000 & 0.6750 & {0.6500} & 0.4750 & 0.6500 & {0.6000} & 0.5750 \\
& Height & 0.5117 & {0.6117} & 0.5883 & 0.6167 & 0.6167 & {0.6517} & 0.5733 \\
& Existence & {0.9750} & {0.9000} & 0.9250 & {0.9750} & 0.9750 & {0.9500} & 0.9000 \\
& Occlusion & 0.5660 & 0.6620 & 0.5880 & 0.6160 & {0.6900} & {0.7940} & 0.6160 \\
& Object Shape & 0.2308 & 0.2846 & 0.2692 & {0.3692} & {0.3154} & 0.3385 & 0.3000 \\
& Counting & {0.5672} & 0.5042 & 0.4454 & 0.3991 & {0.4370} & 0.5462 & 0.4916 \\
& Object Size Estimation & 0.5265 & {0.4679} & 0.5530 & 0.4575 & 0.5590 & 0.4573 & {0.6139} \\
% Spatial Understanding
\SetCell[r=14]{c, font=\bfseries} Spatial Understanding & Spatial Compatibility & 0.5280 & 0.4673 & 0.4813 & 0.4673 & 0.6168 & {0.6075} & {0.5748} \\
& Coordinate Conversion & 0.5838 & {0.5413} & 0.5113 & 0.3563 & {0.6475} & {0.6475} & 0.5900 \\
& Trajectory Description & 0.3205 & {0.5000} & 0.2179 & {0.4615} & 0.3718 & 0.2692 & {0.5513} \\
& Geometric Reasoning & 0.2103 & 0.2738 & {0.2897} & 0.1786 & 0.2659 & {0.2857} & 0.2659 \\
& Spatial Imagination & {0.3100} & 0.3075 & 0.2325 & 0.2575 & 0.2925 & {0.2800} & {0.3100} \\
S & Spatial Grid & 0.4675 & 0.4025 & 0.4375 & 0.5650 & 0.6375 & {0.6425} & {0.7525} \\
& Temporal Appearance Order & 0.1818 & {0.0000} & {0.2727} & {0.3636} & 0.2727 & 0.0909 & {0.2727} \\
& Multi-View Reasoning & 0.3552 & {0.3892} & 0.3287 & {0.4358} & 0.4125 & 0.3934 & 0.3998 \\
& Situational QA & 0.4994 & 0.4972 & 0.4729 & 0.4036 & 0.5329 & {0.5351} & {0.5173} \\
& Velocity Acceleration & 0.3613 & 0.3882 & 0.2605 & 0.3344 & 0.4202 & {0.4840} & {0.5092} \\
& Relative Distance & 0.4695 & {0.4629} & 0.3130 & 0.3608 & 0.5812 & 0.6324 & {0.5462} \\
& Camera Pose & 0.4041 & 0.4477 & 0.3198 & 0.3459 & 0.5029 & {0.5581} & {0.5087} \\
& Spatial Relation & {0.8239} & 0.7370 & 0.5892 & 0.4395 & 0.8130 & {0.8435} & 0.8133 \\
& Object Localization & 0.4704 & {0.7290} & {0.5763} & 0.5670 & 0.3832 & 0.4922 & 0.2461 \\
% Planning
\SetCell[r=2]{c, font=\bfseries} Planning & Maze Navigation & 0.3475 & 0.3375 & 0.2950 & {0.6400} & {0.5650} & 0.4775 & 0.4850 \\
& Route Planning & 0.3220 & 0.3220 & 0.3729 & {0.4068} & 0.3220 & 0.2542 & {0.4576} \\
% Overall
\SetCell[c=2]{c, font=\bfseries} Overall & & 0.4712 & 0.4850 & 0.4278 & 0.4389 & 0.5338 & {0.5481} & {0.5252} \\
\end{tblr}%
}
\end{table*}

% \vspace{-6mm}
\begin{table*}[!t]
  \centering
  \caption{Performance evaluation of different models on the SIBench-mini.}
  \resizebox{\textwidth}{!}{
    \begin{tblr}{
      colspec = {Q[l, m] Q[l, m] *{7}{Q[c, m]}},
      hline{1,Z} = {wd=0.8pt},        % Thick top and bottom lines
      hline{3} = {1-9}{wd=0.4pt},     % Mid-rule under "Models"
      hline{2, 10, 24, 26} = {wd=0.4pt}, % Main section separator lines
    }
    % Header
    \SetCell[r=2, c=2]{c} Settings & & \SetCell[c=7]{c} Models \\
    & & LLaVA-OneVision-72B~\cite{li2024llava} & Qwen2.5-VL-72B~\cite{bai2025qwen2} & InternVL3.5-38B~\cite{zhu2025internvl3} & GLM4.5-V-106B-A12B~\cite{zeng2025glm} & Doubao-Seed-1.6-Vision~\cite{guo2025seed1} & Gemini-2.5-Pro~\cite{team2023gemini} & GPT-5~\cite{gpt5} \\
    % Basic Perception Section
    \SetCell[r=7]{c, font=\bfseries} Basic Perception & Reach Prediction & 0.6000 & {0.6750} & 0.5750 & {0.6500} & 0.5000 & 0.5750 & {0.6000} \\
    & Height & 0.6500 & 0.6000 & 0.5633 & {0.7750} & 0.7500 & 0.7500 & {0.8000} \\
    & Existence & 0.9500 & {0.9500} & 1.000 & {0.9750} & {0.9500} & {0.9750} & 0.9250 \\
    & Occlusion & 0.6000 & 0.5500 & 0.6160 & {0.7250} & {0.8000} & {0.7250} & 0.6250 \\
    & Object Shape & 0.2500 & 0.2750 & 0.3000 & {0.4000} & 0.3250 & 0.2000 & {0.4750} \\
    & Counting & 0.8889 & {0.6111} & 0.4916 & 0.7222 & {0.9444} & 0.8333 & 0.7222 \\
    & Object Size Estimation & 0.5150 & 0.6675 & 0.6139 & 0.5800 & 0.5450 & {0.6175} & {0.6425} \\
    % Spatial Understanding Section
    \SetCell[r=14]{c, font=\bfseries} Spatial Understanding & Spatial Compatibility & 0.6250 & {0.7000} & 0.5748 & 0.7000 & 0.6500 & {0.7250} & 0.6250 \\
    & Coordinate Conversion & 0.7000 & 0.7000 & 0.5900 & {0.8000} & {0.7250} & 0.7000 & 0.6750 \\
    & Trajectory Description & {0.5500} & {0.6250} & {0.5513} & 0.3750 & 0.4500 & 0.4500 & 0.4750 \\
    & Geometric Reasoning & 0.2750 & 0.2500 & 0.2659 & 0.3000 & 0.3000 & {0.4000} & {0.3500} \\
    & Spatial Imagination & 0.3750 & {0.3000} & 0.3100 & {0.3250} & {0.3250} & {0.3250} & {0.3250} \\
    & Spatial Grid & 0.6250 & 0.8250 & 0.7265 & {0.9250} & 0.9750 & {1.0000} & 0.7750 \\
    & Temporal Appearance Order & 0.3636 & 0.5454 & 0.2727 & 0.1818 & {0.9091} & 0.1818 & {1.0000} \\
    & Multi-View Reasoning & 0.5250 & 0.3250 & 0.2269 & 0.3750 & {0.5250} & 0.4500 & {0.5750} \\
    & Situational QA & 0.3750 & 0.3438 & {0.5173} & 0.3438 & 0.5000 & 0.4375 & {0.6250} \\
    & Velocity Acceleration & 0.4000 & 0.4500 & 0.5092 & 0.4750 & {0.5750} & 0.5000 & {0.5250} \\
    & Relative Distance & 0.5325 & 0.5650 & 0.4811 & 0.6775 & {0.7350} & 0.6675 & {0.7075} \\
    & Camera Pose & {0.2250} & 0.4000 & 0.4884 & 0.3000 & 0.4000 & {0.5000} & 0.4000 \\
    & Spatial Relation & 0.6250 & {0.5750} & {0.8133} & 0.7500 & 0.6500 & 0.6750 & 0.6000 \\
    & Object Localization & 0.7250 & 0.7000 & 0.2430 & 0.7500 & {0.8250} & 0.7750 & {0.8000} \\
    % Planning Section
    \SetCell[r=2]{c, font=\bfseries} Planning & Maze Navigation & 0.4000 & 0.5000 & 0.3150 & {0.7750} & {0.8500} & {0.7750} & 0.7500 \\
    & Route Planning & {0.3750} & 0.3000 & {0.4576} & 0.2500 & 0.4500 & 0.3750 & 0.4500 \\
    % Overall Section
    \SetCell[c=2]{c, font=\bfseries} Overall & & 0.5252 & 0.5168 & 0.5355 & 0.5822 & {0.6216} & 0.6295 & {0.6906} \\
    \end{tblr}%
  }
  \label{tab:mini_performance_supp} % 
\end{table*}%
\subsection{Failure case analysis}
\begin{figure}[t]
\centering
\includegraphics[width=3.5in]{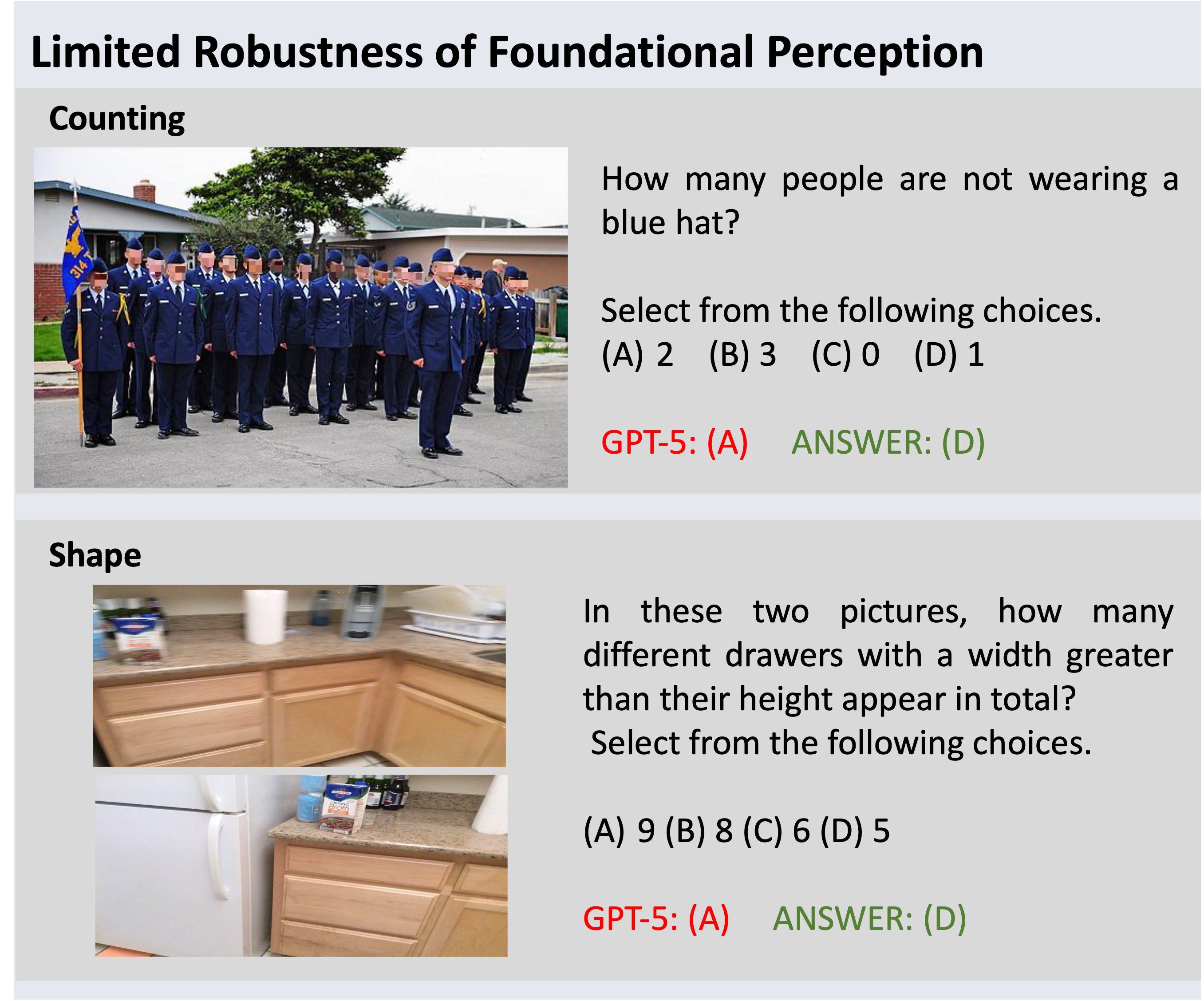}
\caption{\textbf{Limited Robustness of Foundational Perception.}}
\label{fig:perception_challenge}
\end{figure}

As mentioned in \refsection{sec:detail_comparision_of_various_setting}, there are four areas of model weakness. \ie, foundational perception, quantitative capabilities, spatial imagination, and multi-view reasoning. We provide visualizations and analysis of failure cases of these four aspects.
For \textbf{foundational perception}, \reffig{fig:perception_challenge} shows two failure cases, including counting and shape. For counting, GPT-5 predicts close to the GT answer. For shape analysis, GPT-5 predicts only half of GT number of drawers.

\begin{figure}[t]
\centering
\includegraphics[width=3.5in]{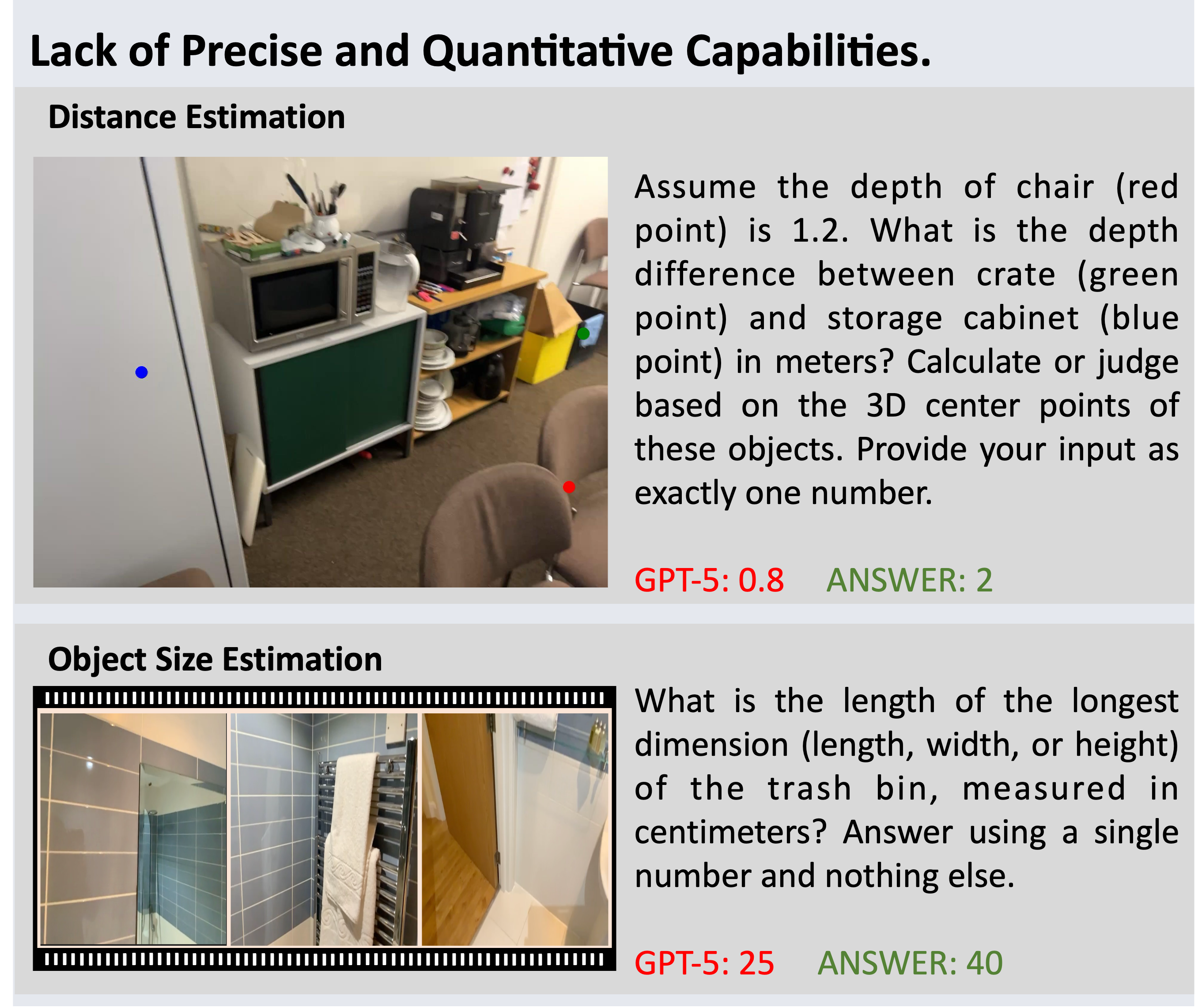}
\caption{\textbf{Lack of Precise and Quantitative Capabilities.}}
\label{fig:quantitative}
\end{figure}

\begin{figure}[t]
\centering
\includegraphics[width=3.5in]{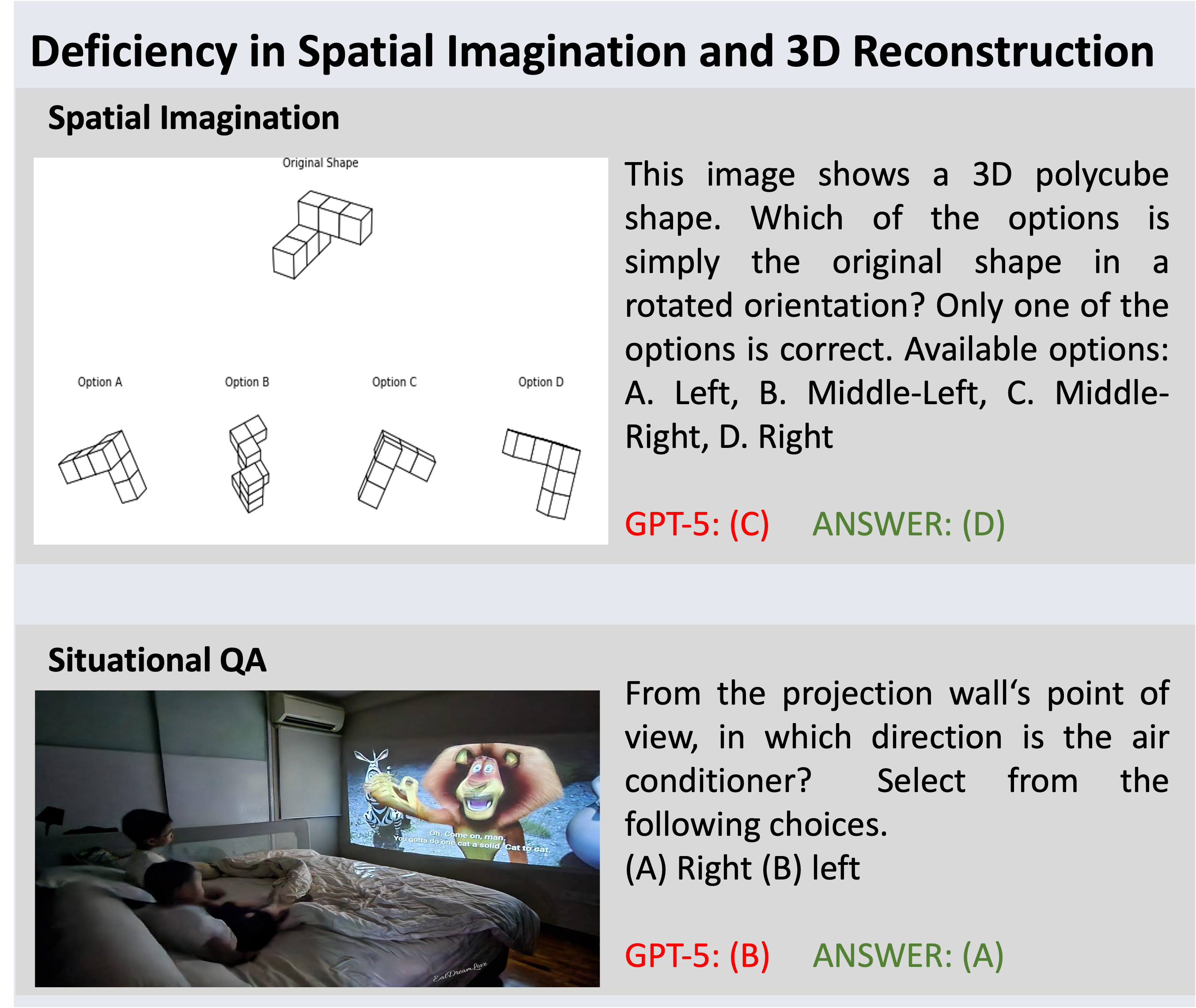}
\caption{\textbf{Deficiency in Spatial Imagination and 3D Reconstruction.}}
\label{fig:imagination}
\end{figure}

\begin{figure}[t]
\centering
\includegraphics[width=3.5in]{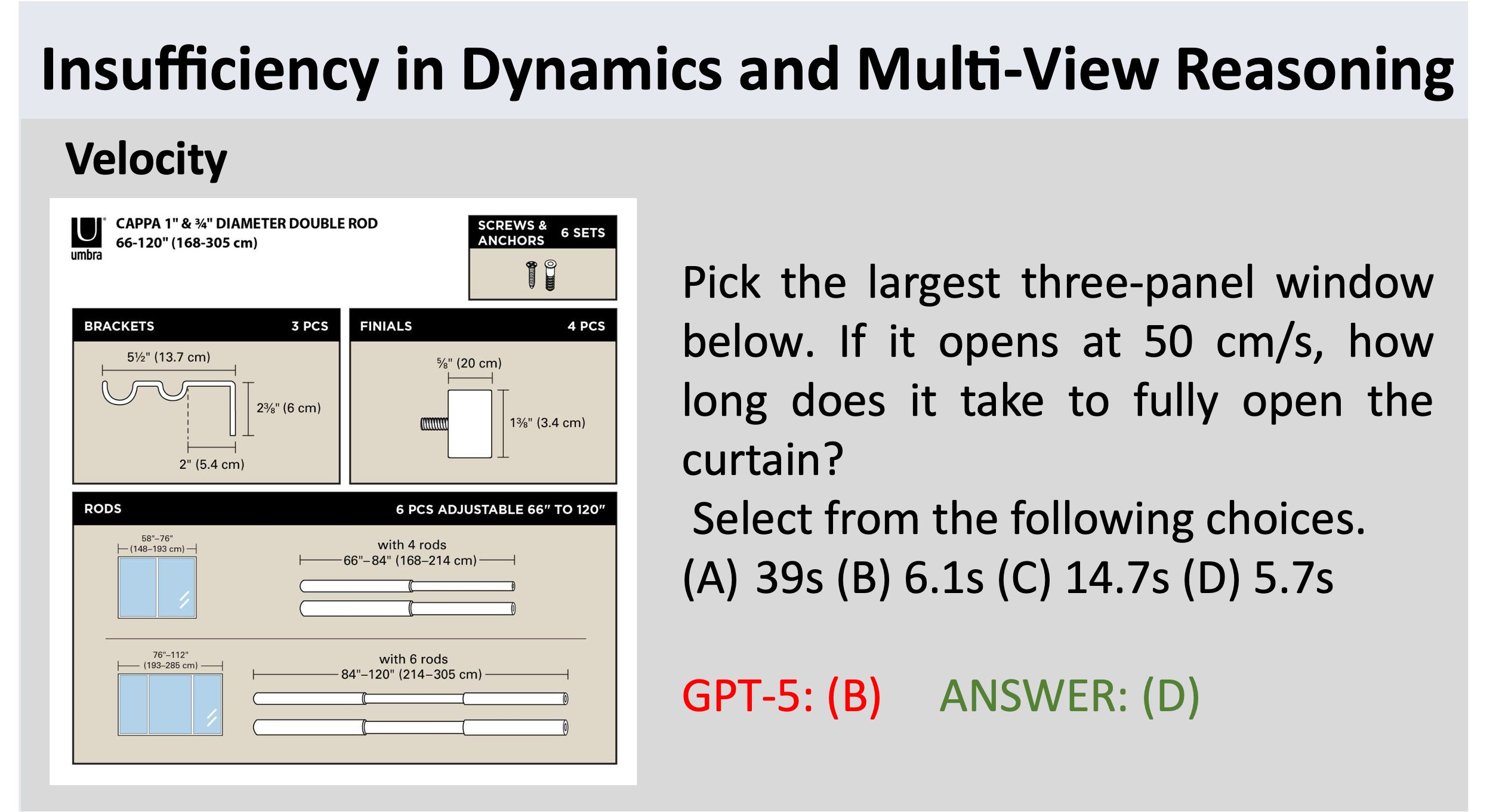}
\caption{\textbf{Insufficiency in Dynamics and Multi-View Reasoning.}}
\label{fig:dynamics}
\end{figure}

For \textbf{precise and quantitative capabilities}, the current VLM also shows limited performance, as shown in \reffig{fig:quantitative}. Specifically, GPT-5 provides answers that significantly deviate from the ground truth when tasked with estimating the distance between objects and the size of a specific object from an image.
Current VLMs also show deficiencies in \textbf{spatial imagination and 3D reconstruction} (See \reffig{fig:imagination}). In these cases, GPT-5 is unable to correctly identify a rotated 3D shape or determine the relative spatial position of an object from a different perspective.
VLMs are weak at \textbf{dynamics and multi-view reasoning}. 
In \reffig{fig:dynamics}, GPT-5 incorrectly answers a question that requires it to first identify the largest window dimension from the provided specifications and then use that data to calculate the time needed to open a curtain at a given velocity.

\subsection{Timeline of Representative VSR Benchmarks}
\begin{figure*}[t]
\centering
\includegraphics[width=7in]{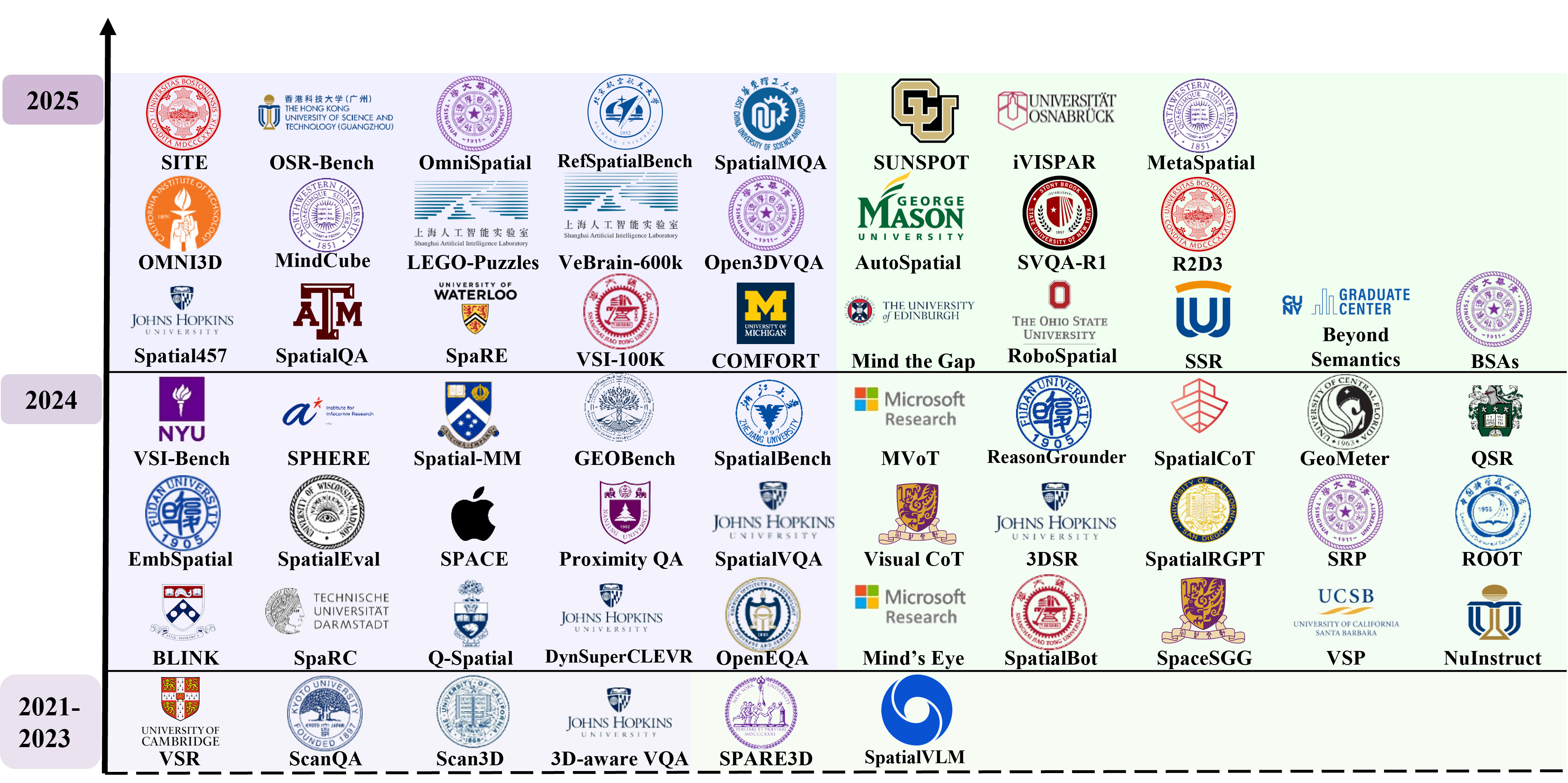}
\caption{\textbf{Timeline of Representative VSR Benchmarks}. Purple indicates novel evaluation methods and green denotes methodological improvements.}
\label{fig:timeline}
\end{figure*}

We survey over 60 benchmarks in VSR and categorize them into two main types of contributions: {methodological advancements and novel benchmarks}, as shown in \reffig{fig:timeline}.
\subsection{Comparison between current benchmarks and SIBench}
\reftab{tab:benchmark_comparison} presents a comparison of current benchmarks with SIBench.
Task settings are categorized as concentrated if there are fewer than 10, and diverse otherwise.
The proposed SIBench is characterized by the extensive coverage of input and QA types, all within a large-scale dataset.

\begin{table*}[!t]
  \centering
  \caption{Comparison of Benchmarks for Visual Spatial Reasoning.}
  \label{tab:benchmark_comparison} % 替换为您想要的标签
  \small % 使用标准小号字体，而不是强制缩放
  \setlength{\tabcolsep}{4pt} % 减少列之间的间距，使表格更紧凑
  % 使用 @{} 去掉表格左右两边的多余空白
  % 使用 l(左对齐), r(右对齐), c(居中对齐) 优化对齐方式
  \begin{tabular}{@{} c c r ccc ccc c @{}}
    \toprule
    \multirow{2.4}{*}{Benchmark} & \multirow{2.4}{*}{Annotation} & \multicolumn{1}{c}{Scale} & \multicolumn{3}{c}{Input Type} & \multicolumn{3}{c}{QA Type} & \multirow{2.4}{*}{Diversity} \\
    \cmidrule(lr){4-6} \cmidrule(lr){7-9} % 使用 (lr) 让分隔线更好看
     & & \multicolumn{1}{c}{(k)} & Single & Multi-View & Video & T/F & Multi-Choice & Number & \\
    \midrule
    VSR~\cite{VSR}            & Synthetic & 2.81      & \textcolor{nvidiagreen}{\ding{51}} & \textcolor{red}{\ding{55}} & \textcolor{red}{\ding{55}} & \textcolor{nvidiagreen}{\ding{51}} & \textcolor{red}{\ding{55}} & \textcolor{red}{\ding{55}} & Concentrated \\
    BLINK~\cite{BLINK}          & Manual    & 0.28      & \textcolor{nvidiagreen}{\ding{51}} & \textcolor{red}{\ding{55}} & \textcolor{red}{\ding{55}} & \textcolor{red}{\ding{55}} & \textcolor{nvidiagreen}{\ding{51}} & \textcolor{red}{\ding{55}} & Diverse \\
    VSI-Bench~\cite{VSI-Bench}     & Synthetic & 5.00      & \textcolor{red}{\ding{55}} & \textcolor{red}{\ding{55}} & \textcolor{nvidiagreen}{\ding{51}} & \textcolor{red}{\ding{55}} & \textcolor{red}{\ding{55}} & \textcolor{nvidiagreen}{\ding{51}} & Diverse \\
    Spatial-MM~\cite{Spatial-MM}     & Synthetic & 2.30      & \textcolor{nvidiagreen}{\ding{51}} & \textcolor{red}{\ding{55}} & \textcolor{red}{\ding{55}} & \textcolor{red}{\ding{55}} & \textcolor{nvidiagreen}{\ding{51}} & \textcolor{red}{\ding{55}} & Concentrated \\
    SpatialRGPT~\cite{SpatialRGPT}    & Synthetic & 1.40      & \textcolor{nvidiagreen}{\ding{51}} & \textcolor{red}{\ding{55}} & \textcolor{red}{\ding{55}} & \textcolor{nvidiagreen}{\ding{51}} & \textcolor{red}{\ding{55}} & \textcolor{nvidiagreen}{\ding{51}} & Concentrated \\
    SpatialVLM~\cite{SpatialVLM}     & Synthetic & 0.54      & \textcolor{nvidiagreen}{\ding{51}} & \textcolor{red}{\ding{55}} & \textcolor{red}{\ding{55}} & \textcolor{red}{\ding{55}} & \textcolor{red}{\ding{55}} & \textcolor{nvidiagreen}{\ding{51}} & Concentrated \\
    STI-Bench~\cite{STI-Bench}      & Synthetic & 2.00      & \textcolor{red}{\ding{55}} & \textcolor{red}{\ding{55}} & \textcolor{nvidiagreen}{\ding{51}} & \textcolor{red}{\ding{55}} & \textcolor{nvidiagreen}{\ding{51}} & \textcolor{red}{\ding{55}} & Diverse \\
    SpatialBench~\cite{SpatialBench}   & Synthetic & 0.39      & \textcolor{nvidiagreen}{\ding{51}} & \textcolor{red}{\ding{55}} & \textcolor{red}{\ding{55}} & \textcolor{nvidiagreen}{\ding{51}} & \textcolor{nvidiagreen}{\ding{51}} & \textcolor{nvidiagreen}{\ding{51}} & Concentrated \\
    RoboSpatial~\cite{Robospatial}    & Synthetic & 6.00   & \textcolor{nvidiagreen}{\ding{51}} & \textcolor{red}{\ding{55}} & \textcolor{red}{\ding{55}} & \textcolor{nvidiagreen}{\ding{51}} & \textcolor{red}{\ding{55}} & \textcolor{nvidiagreen}{\ding{51}} & Concentrated \\
    What's up~\cite{whatsUp}      & Synthetic & 0.49      & \textcolor{nvidiagreen}{\ding{51}} & \textcolor{red}{\ding{55}} & \textcolor{red}{\ding{55}} & \textcolor{red}{\ding{55}} & \textcolor{nvidiagreen}{\ding{51}} & \textcolor{red}{\ding{55}} & Concentrated \\
    Space3D-Bench~\cite{Space3d-Bench}  & Synthetic & 1.00      & \textcolor{nvidiagreen}{\ding{51}} & \textcolor{red}{\ding{55}} & \textcolor{red}{\ding{55}} & \textcolor{red}{\ding{55}} & \textcolor{red}{\ding{55}} & \textcolor{nvidiagreen}{\ding{51}} & Concentrated \\
    EmbSpatial-Bench~\cite{EmbSpatial}& Manual    & 3.64      & \textcolor{nvidiagreen}{\ding{51}} & \textcolor{red}{\ding{55}} & \textcolor{red}{\ding{55}} & \textcolor{red}{\ding{55}} & \textcolor{nvidiagreen}{\ding{51}} & \textcolor{red}{\ding{55}} & Concentrated \\
    OmniSpatial~\cite{OmniSpatial}    & Manual    & 1.50      & \textcolor{nvidiagreen}{\ding{51}} & \textcolor{red}{\ding{55}} & \textcolor{red}{\ding{55}} & \textcolor{nvidiagreen}{\ding{51}} & \textcolor{nvidiagreen}{\ding{51}} & \textcolor{red}{\ding{55}} & Diverse \\
    PhysBench~\cite{PhysBench}      & Manual    & {10.00}     & \textcolor{nvidiagreen}{\ding{51}} & \textcolor{red}{\ding{55}} & \textcolor{nvidiagreen}{\ding{51}} & \textcolor{red}{\ding{55}} & \textcolor{nvidiagreen}{\ding{51}} & \textcolor{red}{\ding{55}} & Concentrated \\
    VSI-100K~\cite{VSI-100K}       & Synthetic & 5.00    & \textcolor{red}{\ding{55}} & \textcolor{red}{\ding{55}} & \textcolor{nvidiagreen}{\ding{51}} & \textcolor{red}{\ding{55}} & \textcolor{red}{\ding{55}} & \textcolor{nvidiagreen}{\ding{51}} & Concentrated \\
    SpatialMQA~\cite{SpatialMQA}     & Manual    & 5.39      & \textcolor{nvidiagreen}{\ding{51}} & \textcolor{red}{\ding{55}} & \textcolor{red}{\ding{55}} & \textcolor{red}{\ding{55}} & \textcolor{nvidiagreen}{\ding{51}} & \textcolor{red}{\ding{55}} & Concentrated \\
    SRBench~\cite{SRBench}        & Synthetic & 1.80      & \textcolor{nvidiagreen}{\ding{51}} & \textcolor{red}{\ding{55}} & \textcolor{red}{\ding{55}} & \textcolor{red}{\ding{55}} & \textcolor{nvidiagreen}{\ding{51}} & \textcolor{red}{\ding{55}} & Concentrated \\
    Open3DVQA~\cite{Open3DVQA}      & Synthetic & 1.24      & \textcolor{red}{\ding{55}} & \textcolor{nvidiagreen}{\ding{51}} & \textcolor{red}{\ding{55}} & \textcolor{nvidiagreen}{\ding{51}} & \textcolor{red}{\ding{55}} & \textcolor{nvidiagreen}{\ding{51}} & Concentrated \\
    Q-Spatial Bench~\cite{Q-Spatial}& Manual    & 0.27      & \textcolor{nvidiagreen}{\ding{51}} & \textcolor{red}{\ding{55}} & \textcolor{red}{\ding{55}} & \textcolor{red}{\ding{55}} & \textcolor{red}{\ding{55}} & \textcolor{nvidiagreen}{\ding{51}} & Concentrated \\
    % Visual-CoT     & Synthetic & N/A    & \textcolor{nvidiagreen}{\ding{51}} & \textcolor{red}{\ding{55}} & \textcolor{red}{\ding{55}} & \textcolor{nvidiagreen}{\ding{51}} & \textcolor{nvidiagreen}{\ding{51}} & \textcolor{red}{\ding{55}} & Concentrated \\
    VoT~\cite{vot}            & Synthetic & 3.51      & \textcolor{nvidiagreen}{\ding{51}} & \textcolor{red}{\ding{55}} & \textcolor{red}{\ding{55}} & \textcolor{red}{\ding{55}} & \textcolor{nvidiagreen}{\ding{51}} & \textcolor{red}{\ding{55}} & Concentrated \\
    % SpatialEval    & Synthetic & {N/A}     & \textcolor{nvidiagreen}{\ding{51}} & \textcolor{red}{\ding{55}} & \textcolor{red}{\ding{55}} & \textcolor{red}{\ding{55}} & \textcolor{nvidiagreen}{\ding{51}} & \textcolor{red}{\ding{55}} & Concentrated \\
    3DSRBench~\cite{3dsrbench}      & Synthetic & 2.77      & \textcolor{nvidiagreen}{\ding{51}} & \textcolor{red}{\ding{55}} & \textcolor{red}{\ding{55}} & \textcolor{nvidiagreen}{\ding{51}} & \textcolor{nvidiagreen}{\ding{51}} & \textcolor{red}{\ding{55}} & Diverse \\
    Super-CLEVR-3D~\cite{Super-CLEVR-3D} & Synthetic & 5.00    & \textcolor{nvidiagreen}{\ding{51}} & \textcolor{red}{\ding{55}} & \textcolor{red}{\ding{55}} & \textcolor{nvidiagreen}{\ding{51}} & \textcolor{nvidiagreen}{\ding{51}} & \textcolor{red}{\ding{55}} & Concentrated \\
    SPHERE~\cite{SPHERE}         & Manual & 2.28      & \textcolor{nvidiagreen}{\ding{51}} & \textcolor{red}{\ding{55}} & \textcolor{red}{\ding{55}} & \textcolor{nvidiagreen}{\ding{51}} & \textcolor{red}{\ding{55}} & \textcolor{nvidiagreen}{\ding{51}} & Diverse \\
    LEGO-Puzzles~\cite{LEGO-Puzzles}   & Manual    & 1.10      & \textcolor{nvidiagreen}{\ding{51}} & \textcolor{red}{\ding{55}} & \textcolor{nvidiagreen}{\ding{51}} & \textcolor{nvidiagreen}{\ding{51}} & \textcolor{nvidiagreen}{\ding{51}} & \textcolor{nvidiagreen}{\ding{51}} & Diverse \\
    GEOBench-VLM~\cite{GeoBench}   & Manual    & {10.00}     & \textcolor{nvidiagreen}{\ding{51}} & \textcolor{red}{\ding{55}} & \textcolor{nvidiagreen}{\ding{51}} & \textcolor{red}{\ding{55}} & \textcolor{nvidiagreen}{\ding{51}} & \textcolor{red}{\ding{55}} & Diverse \\
    OpenEQA~\cite{OpenEQA}        & Manual    & 1.60      & \textcolor{red}{\ding{55}} & \textcolor{nvidiagreen}{\ding{51}} & \textcolor{red}{\ding{55}} & \textcolor{red}{\ding{55}} & \textcolor{red}{\ding{55}} & \textcolor{nvidiagreen}{\ding{51}} & Concentrated \\
    % SPACE          & Synthetic & {N/A}     & \textcolor{nvidiagreen}{\ding{51}} & \textcolor{nvidiagreen}{\ding{51}} & \textcolor{nvidiagreen}{\ding{51}} & \textcolor{nvidiagreen}{\ding{51}} & \textcolor{nvidiagreen}{\ding{51}} & \textcolor{nvidiagreen}{\ding{51}} & Diverse \\
    SPARE3D~\cite{SPARE3D}        & Synthetic & 2.50     & \textcolor{red}{\ding{55}} & \textcolor{nvidiagreen}{\ding{51}} & \textcolor{red}{\ding{55}} & \textcolor{red}{\ding{55}} & \textcolor{nvidiagreen}{\ding{51}} & \textcolor{red}{\ding{55}} & Concentrated \\
    VSP~\cite{VSP} & Synthetic & 4.40      & \textcolor{nvidiagreen}{\ding{51}} & \textcolor{red}{\ding{55}} & \textcolor{red}{\ding{55}} & \textcolor{red}{\ding{55}} & \textcolor{nvidiagreen}{\ding{51}} & \textcolor{red}{\ding{55}} & Diverse \\
    DynSuperCLEVR~\cite{DynSuperCLEVR}  & Synthetic & 1.00     & \textcolor{red}{\ding{55}} & \textcolor{red}{\ding{55}} & \textcolor{nvidiagreen}{\ding{51}} & \textcolor{nvidiagreen}{\ding{51}} & \textcolor{nvidiagreen}{\ding{51}} & \textcolor{red}{\ding{55}} & Concentrated \\
    % EmbSpatial-Bench~\cite{EmbSpatial} & Manual    & 3.64      & \textcolor{red}{\ding{55}} & \textcolor{nvidiagreen}{\ding{51}} & \textcolor{red}{\ding{55}} & \textcolor{red}{\ding{55}} & \textcolor{nvidiagreen}{\ding{51}} & \textcolor{red}{\ding{55}} & Concentrated \\
    SITE ~\cite{SITE}           & Synthetic & 8.06      & \textcolor{nvidiagreen}{\ding{51}} & \textcolor{red}{\ding{55}} & \textcolor{nvidiagreen}{\ding{51}} & \textcolor{red}{\ding{55}} & \textcolor{nvidiagreen}{\ding{51}} & \textcolor{red}{\ding{55}} & Diverse \\
    \midrule
    {SIBench}  & {Manual}    & 9.00      & \textcolor{nvidiagreen}{\ding{51}} & \textcolor{nvidiagreen}{\ding{51}} & \textcolor{nvidiagreen}{\ding{51}} & \textcolor{nvidiagreen}{\ding{51}} & \textcolor{nvidiagreen}{\ding{51}} & \textcolor{nvidiagreen}{\ding{51}} & {Diverse} \\
    \bottomrule
  \end{tabular}
\end{table*} 
% 示例：您可以在这里继续添加
% \section{More Appendix}
% ...
\end{appendices}
% --- 附录结束 ---

\vfill

% that's all folks
\end{document}